\documentclass[10pt,twocolumn,letterpaper]{article}

\usepackage[pagenumbers]{cvpr}

\newcommand{\hermes}{HERMES\xspace}
\usepackage{multirow}
\usepackage{colortbl}
\definecolor{CapGen}{HTML}{DCEBF8}\definecolor{CapMed}{HTML}{DDF0E1}\definecolor{CapPath}{HTML}{FADDE0}\definecolor{CapOurs}{HTML}{E9DDF6}
\usepackage{array}
\newcommand{\cmark}{{\color{ForestGreen}\ding{51}}}
\newcommand{\xmark}{{\color{BrickRed}\ding{55}}}
\newcommand{\pmark}{{\color{BurntOrange}$\triangle$}}
\usepackage{pifont}

\usepackage{tabularx}

\definecolor{cvprblue}{rgb}{0.21,0.49,0.74}
\usepackage[pagebackref,breaklinks,colorlinks,allcolors=cvprblue]{hyperref}

\def\confName{}
\def\confYear{}

\title{Beyond Final Answers: Interpretable Tumor Immune Microenvironment Reasoning from H\&E and mIF}

\author{Yixuan Duan$^{1}$ \qquad Kexin Geng$^{1}$ \qquad Hanwen Xu$^{2,\dagger}$ \qquad Wei Qiu$^{1,\dagger}$\\[4pt]
$^{1}$Rice University \qquad $^{2}$University of Pennsylvania\\[2pt]
{\small $^{\dagger}$Co-last authors}
}

\begin{document}
\twocolumn[{%
\renewcommand\twocolumn[1][]{#1}%
\maketitle
\begin{center}
  \includegraphics[width=\linewidth]{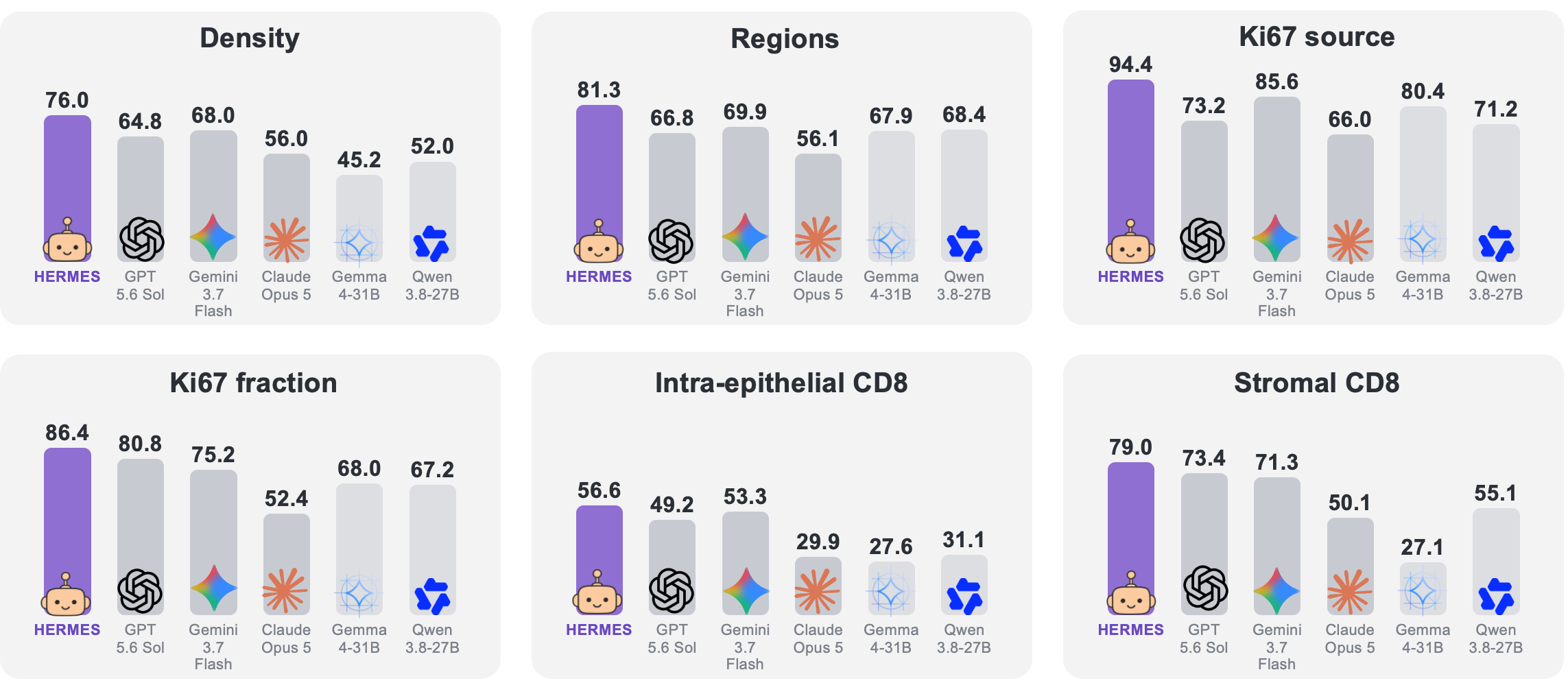}
  \captionof{figure}{\hermes (4B) vs.\ leading MLLMs on held-out Orion patients. Density, Regions and Ki67: accuracy (\%); CD8: clinical-grade agreement ($\kappa \times 100$). Full results in \cref{tab:main}.}
  \label{fig:teaser}
\end{center}
\vspace{1em}
}]
\begin{abstract}
Readouts of the tumor immune microenvironment (TIME), such as CD8$^+$ T-cell density, intra-epithelial infiltration and the Ki67 index, combine tissue architecture from H\&E with cell phenotypes from multiplex immunofluorescence (mIF).
Multimodal large language models (MLLMs) could automate these readouts with an explanation, but two gaps remain.
Pathology MLLMs read H\&E alone, while these readouts need the cell phenotypes that mIF measures. And their reasoning has no step-level truth to be checked against, so R1-style reinforcement learning rewards only the final answer and leaves the reasoning unverified.
Same-section H\&E and mIF provide what both gaps lack: mIF measures every cell without manual annotation, which gives cell-level ground truth for each step that clinical guidelines prescribe, from delineating a tissue compartment to counting and normalising the marker-positive cells within it.
We therefore present \hermes (\textbf{H}\&\textbf{E}--mIF \textbf{R}easoning with \textbf{M}ultimodal \textbf{E}vidence \textbf{S}upervision), an R1-style MLLM that reasons jointly over paired H\&E and mIF in guideline steps, each naming the modality it reads, and is trained with a process reward that checks every stated quantity against this truth.
Across four field-level TIME tasks in colorectal cancer, the 4B-parameter \hermes reaches an average accuracy of 84.5\%, compared with 74.7\% for the strongest of twelve proprietary, general, medical and pathology MLLMs, and a higher share of its intermediate quantities is correct than for the strongest open-weight baseline.
The process reward is what makes the stated reasoning correct: without it, reinforcement learning barely improves on supervised fine-tuning (80.0\% vs. 78.8\%) and fewer stated quantities are correct, and the answers follow from the stated quantities, changing as implied in all 885 test questions when a decisive quantity is edited.
Without retraining, it keeps the highest average accuracy on an external cohort imaged with CODEX (76.4\% vs.\ 67.3\%) and on head and neck cancer, a cancer type absent from training (76.7\% vs.\ 72.5\%).
\end{abstract}
    
\section{Introduction}
\label{sec:intro}

\begin{figure*}[t]
  \centering
  \includegraphics[width=\linewidth]{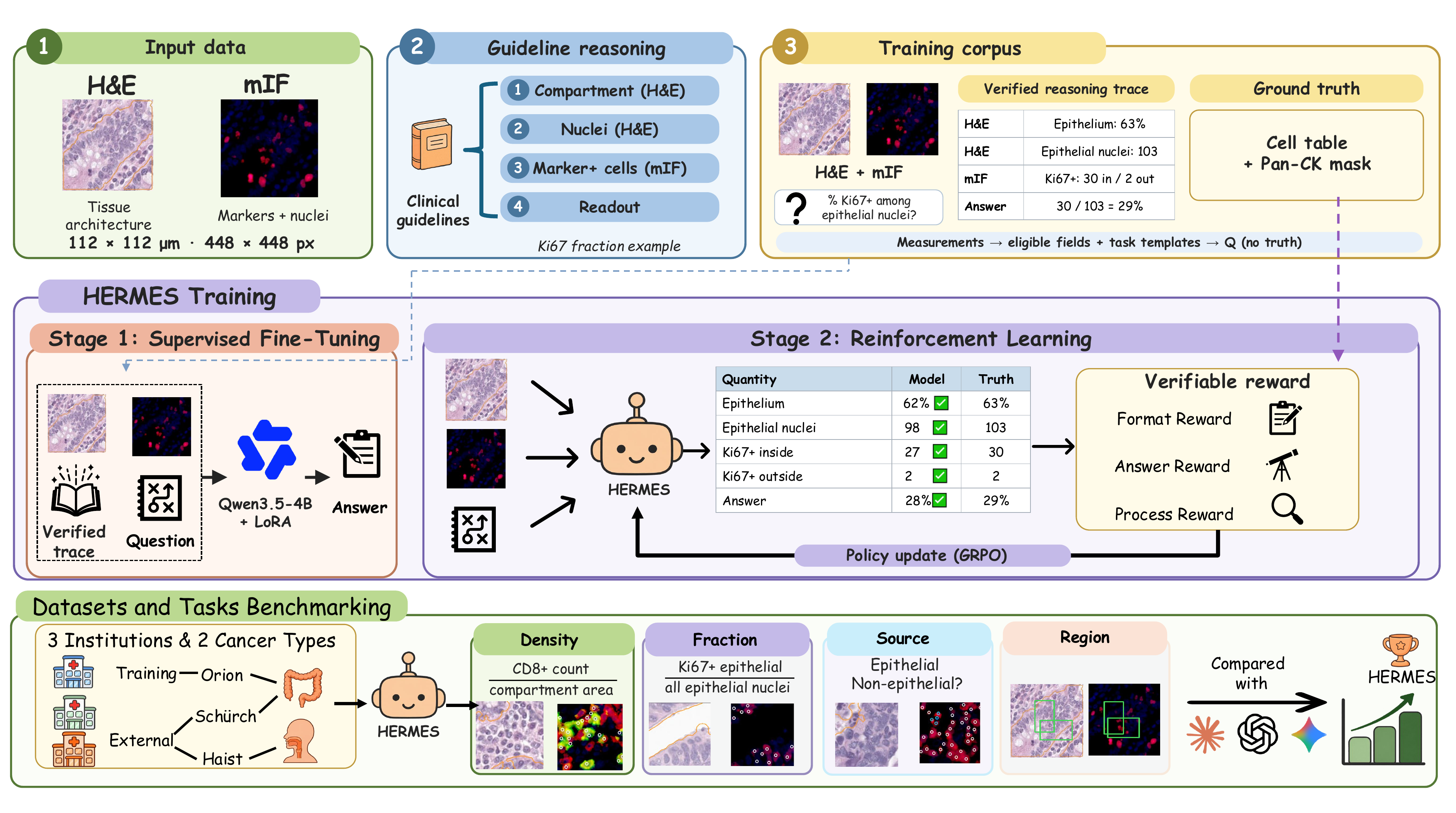}
  \caption{Overview of \hermes. (1) A same-section H\&E and mIF field. (2) Guideline-structured reasoning: compartment and nuclei on H\&E, marker-positive cells on mIF, then the readout. (3) Training traces are built from the cell-level ground truth (cell table, epithelial Pan-CK mask), which the model never sees. Stage 1: SFT of Qwen3.5-4B with LoRA; stage 2: GRPO with format, answer and process rewards, the last checking each stated quantity against the truth (\checkmark: within tolerance). Bottom: four tasks, evaluated on Orion and two external cohorts against proprietary and open-weight MLLMs.}
  \label{fig:method}
\end{figure*}

Characterizing the tumor immune microenvironment (TIME) is a fundamental task in computational pathology, as it shapes how colorectal cancers progress and respond to therapy~\cite{binnewies2018time,fridman2012immune}.
The readouts used in practice quantify specific cells in specific tissue compartments: the Immunoscore is built on the densities of CD3$^+$ and CD8$^+$ T cells in the tumor and at its margin~\cite{galon2006type,pages2018immunoscore}, tumor-infiltrating lymphocytes are scored as a share of the stromal area~\cite{salgado2015tils,hendry2017tils}, and the Ki67 index is the share of \emph{tumor} nuclei that are Ki67-positive~\cite{dowsett2011ki67,nielsen2021ki67}.
Location matters as much as number: the same count of cytotoxic T cells indicates an inflamed tumor when they enter the epithelium and an immune-excluded one when they are held back in the stroma~\cite{galon2006type,chen2017elements}.

While H\&E provides the tissue architecture (where the epithelium ends and the stroma begins), multiplex immunofluorescence (mIF) provides the protein identity of each cell (which cells are CD8$^+$ or Ki67$^+$).
The two modalities offer complementary views of the same tissue, and neither suffices on its own: H\&E cannot tell a CD8$^+$ T cell from other lymphocytes, and mIF without an epithelial marker cannot tell a tumor nucleus from a stromal one.
Platforms that image H\&E and mIF on the \emph{same} section now produce such paired data at scale~\cite{lewis2021spatial,lin2023orion,schurch2020coordinated}.
Multimodal large language models (MLLMs) could turn these data into clinical readouts and explain their answers in language a pathologist can follow.
However, building such a model presents two key challenges:

\noindent\textbf{1. Reasoning jointly over H\&E and mIF.}
General and medical~\cite{liu2023llava,tu2024medpalmm,sellergren2025medgemma} and pathology MLLMs~\cite{sun2024pathasst,seyfioglu2024quiltllava,sun2025pathgen,zhang2025pathor1} are trained on H\&E or natural images and have not learned to read mIF together with H\&E.
We therefore structure the reasoning after clinical scoring guidelines: each step names the modality it reads and states one quantity, with the compartment and nuclei taken from H\&E, the marker-positive cells from mIF, and the readout computed from them.

\noindent\textbf{2. Supervising the reasoning, not only the answer.}
Stated reasoning need not reflect how an answer was produced~\cite{turpin2023unfaithful,lanham2023faithfulness}, and R1-style reinforcement learning~\cite{guo2025deepseekr1,shao2024deepseekmath}, which has improved medical MLLMs~\cite{lai2025medr1,pan2025medvlmr1,zhang2025pathor1}, rewards only the final answer.
In our experiments, the strongest open-weight baseline states only about a third of the quantities a question requires, and only 16\% of the required quantities are correct.
Process supervision has been effective for mathematical reasoning~\cite{uesato2022process,lightman2024verify,wang2024mathshepherd}, but needs the truth of every step, which medical VQA has lacked.
Same-section imaging provides exactly this: because every cell is measured in both modalities, cell tables and a Pan-CK epithelial mask give a measured value for every quantity a pathologist would state on the way to an answer.
We hypothesize that rewarding each stated quantity against this cell-level truth makes the reasoning a target that can be learned and verified.

To this end, we introduce \hermes (\textbf{H}\&\textbf{E}--mIF \textbf{R}easoning with \textbf{M}ultimodal \textbf{E}vidence \textbf{S}upervision) (\cref{fig:method}).
\hermes receives an H\&E image and a marker-specific mIF image of the same field and answers four questions that mirror clinical readouts and cover quantitative (CD8$^+$ density in a compartment; Ki67 index), localization (boxing intra-epithelial CD8$^+$ infiltrates) and attribution (whether Ki67$^+$ cells are mainly epithelial) reasoning.
We first fine-tune a 4B-parameter MLLM on guideline-structured reasoning traces built from the measurements, then train it with GRPO~\cite{shao2024deepseekmath} using a reward for the final answer plus a process reward that checks each stated quantity against the cell-level truth, under a tolerance that tightens during training.
On held-out patients of the Orion cohort~\cite{lin2023orion}, \hermes reaches 84.5\% average accuracy, compared with at most 74.7\% for twelve proprietary, general, medical and pathology MLLMs (\cref{fig:teaser}), and 66\% of its intermediate quantities are correct, compared with 16\% for the strongest open-weight baseline. Without the process reward, the same reinforcement learning barely improves on supervised fine-tuning (80.0\% vs.\ 78.8\%), and fewer stated quantities are correct.

\begin{table}[t]
\centering\footnotesize
\setlength{\tabcolsep}{2.5pt}\renewcommand{\arraystretch}{1.15}
\resizebox{\linewidth}{!}{%
\begin{tabular}{@{}lcccc@{}}
\toprule
Trained with & \begin{tabular}[c]{@{}c@{}}General\\MLLM\end{tabular} & \begin{tabular}[c]{@{}c@{}}Medical\\MLLM\end{tabular} & \begin{tabular}[c]{@{}c@{}}Pathology\\MLLM\end{tabular} & \hermes \\
\midrule
Pathology-specific data                & \xmark & \pmark & \cmark & \cmark \\
mIF images                             & \xmark & \xmark & \xmark & \cmark \\
Same-section H\&E + mIF                & \xmark & \xmark & \xmark & \cmark \\
Guideline-structured reasoning         & \xmark & \xmark & \xmark & \cmark \\
Reward on intermediate quantities      & \xmark & \xmark & \xmark & \cmark \\
\bottomrule
\end{tabular}}
\caption{Training comparison of MLLM families (models in \cref{tab:main}). \cmark: yes; \xmark: no or not reported; \pmark: pathology among other medical domains.}
\label{tab:capability}
\end{table}

To summarize, our contributions are
(1) guideline-structured reasoning for TIME readouts, which turns three scoring rules of clinical practice, the compartment area as the denominator~\cite{salgado2015tils,hendry2017tils}, CD8$^+$ cells as a density~\cite{galon2006type,pages2018immunoscore} and the Ki67 index over tumor nuclei only~\cite{dowsett2011ki67,nielsen2021ki67}, into a fixed sequence of steps, each reading one modality and stating one quantity, so that a pathologist can audit every step and same-section H\&E and mIF give every step a measurable truth;
(2) \hermes, an MLLM trained with reinforcement learning for TIME interpretation from paired H\&E and mIF, which learns this structure from traces filled with the cell-level truth and is trained with a process reward that verifies every stated quantity (\cref{tab:capability});
(3) the first reasoning benchmark on same-section H\&E and mIF, with four tasks tied to clinical readouts, cell-level truth for every intermediate quantity, and external cohorts from a different imaging platform and a different cancer type;
(4) a series of experiments showing higher answer accuracy and clinical-grade agreement than twelve MLLMs and the highest accuracy without retraining on a CODEX cohort~\cite{schurch2020coordinated} (76.4\% vs.\ 67.3\%) and on head and neck cancer, with about 190 output tokens against 1.2k--6.5k for the strongest open-weight baselines;
and (5) a multi-level verification of the reasoning: each stated quantity is checked against the cell-level truth, answer accuracy rises with their correctness, editing a stated quantity changes the answer as implied, and removing CD8$^+$ cells from the mIF image lowers the stated count.
\section{Related Work}
\label{sec:related}

\paragraph{Vision-language models for pathology.}
Contrastive and foundation models learn strong H\&E representations (PLIP~\cite{huang2023plip}, Quilt-1M~\cite{ikezogwo2023quilt}, CONCH~\cite{lu2024conch}, UNI~\cite{chen2024uni}, Prov-GigaPath~\cite{xu2024gigapath}, Virchow~\cite{vorontsov2024virchow}), and generative assistants answer free-form questions on them (LLaVA-Med~\cite{li2023llavamed}, Med-PaLM M~\cite{tu2024medpalmm}, HuatuoGPT-Vision~\cite{chen2024huatuogptvision}, PathAsst~\cite{sun2024pathasst}, PathChat~\cite{lu2024pathchat}, Quilt-LLaVA~\cite{seyfioglu2024quiltllava}, PathGen-LLaVA~\cite{sun2025pathgen}, MedGemma~\cite{sellergren2025medgemma}, SlideChat~\cite{chen2025slidechat}), mostly built on the LLaVA recipe~\cite{liu2023llava}.
Trained on H\&E and human captions, they neither read mIF nor state quantities that can be checked against the tissue; \hermes reads both images and states every intermediate quantity.

\paragraph{Reinforcement learning for multimodal reasoning.}
Chain-of-thought prompting~\cite{wei2022cot}, RL from human feedback~\cite{ouyang2022instructgpt} and RL with verifiable rewards (DeepSeek-R1~\cite{guo2025deepseekr1}, GRPO~\cite{shao2024deepseekmath}) improve reasoning, and have been carried over to perception (Visual-RFT~\cite{liu2025visualrft}, Vision-R1~\cite{huang2025visionr1}) and medicine (Med-R1~\cite{lai2025medr1}, MedVLM-R1~\cite{pan2025medvlmr1}, Patho-R1~\cite{zhang2025pathor1}).
Their rewards check only format and final answer.
Process supervision~\cite{uesato2022process,lightman2024verify,wang2024mathshepherd} constrains the steps but needs human step labels or a learned reward model; same-section H\&E and mIF give the truth of every step directly, so \hermes receives an exact process reward without either.

\paragraph{Histology and spatial proteomics.}
Multiplexed imaging (CyCIF~\cite{lin2018cycif}, CODEX~\cite{goltsev2018codex}, IMC~\cite{giesen2014imc}, MIBI~\cite{keren2018mibi}; review in~\cite{lewis2021spatial}) measures tens of proteins per cell, turned into cell tables by segmentation~\cite{greenwald2022mesmer,graham2019hovernet}, and datasets now pair it with H\&E of the same section (Orion~\cite{lin2023orion}) or provide large CODEX cohorts~\cite{schurch2020coordinated}.
Prior work predicts markers from H\&E, following virtual staining~\cite{rivenson2019virtual} (MIPHEI~\cite{balezo2025miphei}, GigaTIME~\cite{valanarasu2025gigatime}), or predicts single readouts from H\&E such as TIL maps~\cite{saltz2018tils}, microsatellite instability~\cite{kather2019msi} or cancer presence~\cite{campanella2019clinical}, without an auditable rationale.
We use measured mIF as input and the cell tables as supervision for reasoning towards clinical readouts: CD8 density as in the Immunoscore~\cite{galon2006type,pages2018immunoscore}, compartment-resolved TILs~\cite{salgado2015tils,hendry2017tils} and the Ki67 index~\cite{nielsen2021ki67}.

\paragraph{Faithful and verifiable reasoning.}
Stated chains of thought need not reflect how a model reached its answer~\cite{jacovi2020faithful,turpin2023unfaithful,lanham2023faithfulness}, and faithfulness is usually probed indirectly or enforced by executing the chain~\cite{lyu2023faithfulcot}; in high-stakes settings, interpretable reasoning is preferable to post-hoc explanation~\cite{rudin2019stop}.

\section{Method}
\label{sec:method}

\begin{table*}[t]
\centering\small
\setlength{\tabcolsep}{3.5pt}
\providecommand{\hd}[1]{\begin{tabular}[b]{@{}c@{}}#1\end{tabular}}
\caption{External cohorts without retraining: Sch\"urch (colorectal cancer, CODEX) and, as a new cancer type, head and neck squamous cell carcinoma (HNSCC, Haist \etal~\cite{haist2026lymph}). Metrics as in \cref{tab:main}.}
\label{tab:schurch}
\begin{tabular}{@{}l cccccc ccc c@{}}
\toprule
 & \multicolumn{5}{c}{Answer accuracy} & & \multicolumn{3}{c}{Clinical grade agreement} & New cancer \\
\cmidrule(lr){2-6}\cmidrule(lr){8-10}\cmidrule(lr){11-11}
Model & Density & Regions & \hd{Ki67\\source} & \hd{Ki67\\fraction} & Average & \hd{Macro-\\F1} & Ki67 & \hd{Intra-epithelial\\CD8} & \hd{Stromal\\CD8} & \hd{HNSCC\\average} \\
\midrule
\rowcolor{gray!12}\multicolumn{11}{c}{\emph{Proprietary MLLMs}} \\
Gemini 3.7 Flash & 66.4 & 58.1 & 82.8 & 62.0 & 67.3 & 68.1 & 77.3 & 32.1 & \textbf{58.9} & 72.5 \\
\rowcolor{gray!12}\multicolumn{11}{c}{\emph{General MLLMs}} \\
Gemma-4-31B & 38.8 & 60.5 & 79.2 & 54.0 & 58.1 & 62.1 & 71.9 & 4.0 & 32.9 & 64.5 \\
Qwen3.8-27B & 58.8 & 59.7 & 72.0 & 56.0 & 61.6 & 62.4 & 67.8 & 29.6 & 33.1 & 59.8 \\
Ministral-3-14B & 31.2 & 51.6 & 65.6 & 27.6 & 44.0 & 44.2 & 17.7 & 6.1 & 6.1 & 34.0 \\
\rowcolor{gray!12}\multicolumn{11}{c}{\emph{Medical MLLMs}} \\
MedGemma-27B & 15.2 & 54.0 & 46.8 & 21.2 & 34.3 & 32.6 & 4.9 & 1.6 & 0.7 & 36.9 \\
HuatuoGPT-3-27B & 56.0 & 59.7 & 66.0 & 51.6 & 58.3 & 60.7 & 69.9 & 23.1 & 48.2 & 59.0 \\
\rowcolor{gray!12}\multicolumn{11}{c}{\emph{Pathology MLLMs}} \\
Quilt-LLaVA-7B & 5.2 & 50.0 & 5.6 & 3.6 & 16.1 & 34.6 & -1.7 & -7.2 & 2.4 & 14.1 \\
PathGen-LLaVA-13B & 2.0 & 50.8 & 0.0 & 0.0 & 13.2 & 27.9 & -29.4 & -13.4 & -3.1 & 12.6 \\
Patho-R1-7B & 47.6 & 48.4 & 28.4 & 19.2 & 35.9 & 36.6 & -6.3 & -7.5 & 0.2 & 35.9 \\
\midrule
Qwen3.5-4B (base) & 27.2 & 58.1 & 61.2 & 36.4 & 45.7 & 52.8 & 53.8 & 10.1 & 5.6 & 44.4 \\
\hermes (SFT) & 54.4 & 64.5 & 90.0 & 74.8 & 70.9 & 69.6 & 84.8 & 42.2 & 55.1 & 73.7 \\
\hermes (RL) & \textbf{69.6} & \textbf{67.7} & \textbf{92.4} & \textbf{76.0} & \textbf{76.4} & \textbf{75.3} & \textbf{87.0} & \textbf{43.9} & \textbf{58.9} & \textbf{76.7} \\
\bottomrule
\end{tabular}
\end{table*}

\subsection{Data and ground truth}
\label{sec:data}
\begin{figure*}[t]
  \centering
  \includegraphics[width=\linewidth]{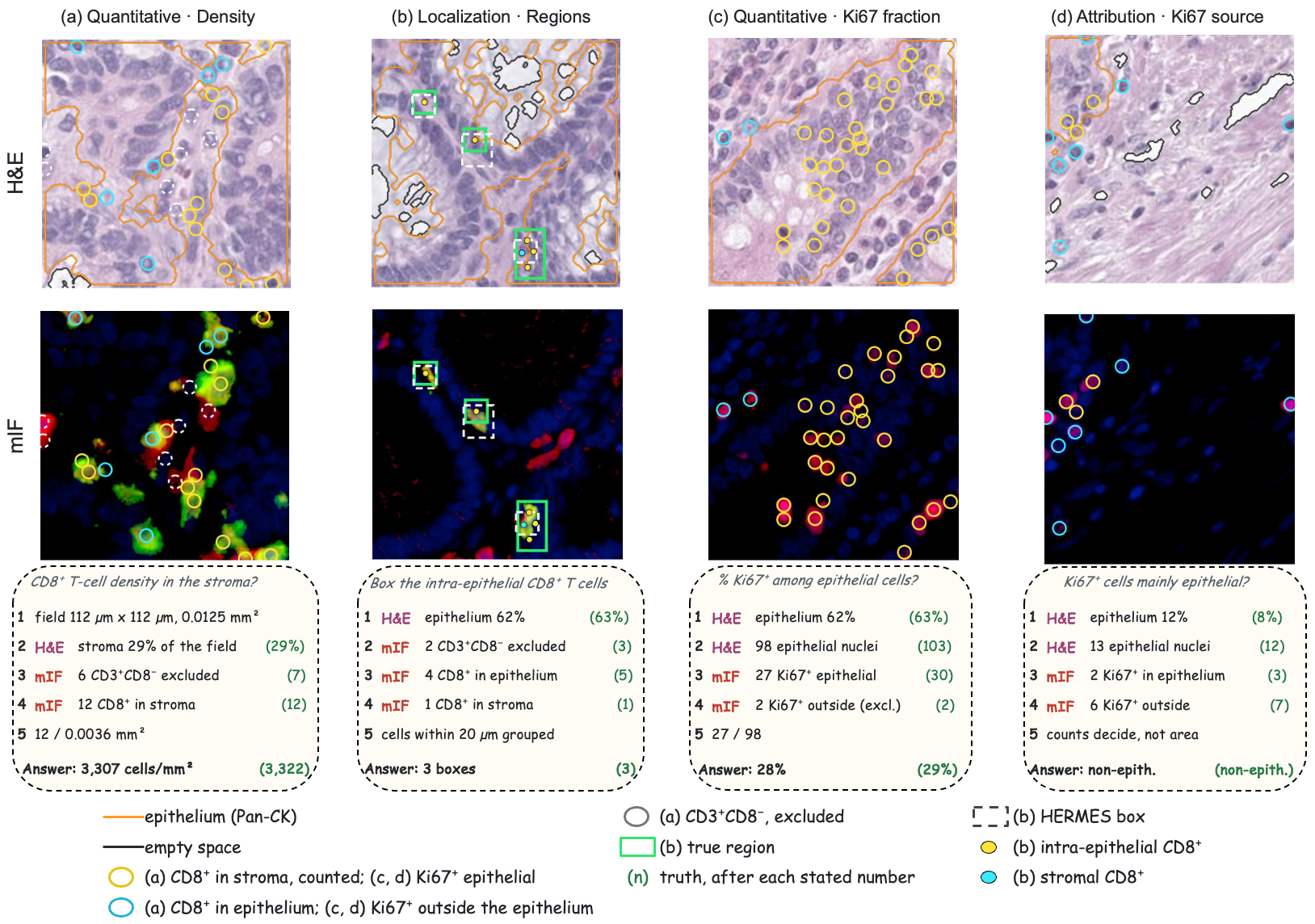}
  \caption{Ground truth and \hermes's reasoning on one held-out field for each of the four tasks: quantitative (a, c), localization (b) and attribution (d). Top: H\&E; middle: mIF; bottom: \hermes's reasoning, with the truth in brackets.}
  \label{fig:gt}
\end{figure*}
\paragraph{Cohorts and fields.}
We use the Orion colorectal cancer cohort~\cite{lin2023orion}, in which each whole-slide section is imaged with cyclic immunofluorescence and then H\&E-stained and scanned, so that both modalities show the same cells.
Patients are split 30/5/5 into training, validation and test sets.
Each question refers to one field of $112\times112\,\mu$m (0.0125\,mm$^2$).
The model receives two images of the field at $448\times448$ pixels: the H\&E crop and a pseudo-coloured mIF image that shows only the markers the question needs (CD3e in red, CD8a in green and Hoechst in blue for the immune tasks, so that CD8$^+$ T cells appear yellow; Ki67 in red and Hoechst in blue for the proliferation tasks).
Each marker is rendered with a fixed gain calibrated on training fields.
No epithelial marker is shown, so the model has to find the epithelium in the H\&E image.

\paragraph{Ground truth.}
All truth is computed from the cell table and the mIF channels and a tissue mask from H\&E.
The epithelium, which holds the tumor cells in these carcinomas, is a mask derived from the Pan-CK channel (smoothed Pan-CK signal, joined with small discs around Pan-CK$^+$ cells and cleaned morphologically).
Tissue is detected from H\&E; empty space is the area that is neither tissue nor epithelium, and stroma is the remaining tissue.
A cell is epithelial if its nucleus lies inside the mask.
A cell is marker-positive if it is called positive in the cell table (per-marker mixture-model calls, as released with~\cite{balezo2025miphei}) \emph{and} is visible in the rendered image, so that every positive cell in the truth can be seen by the model; a CD8$^+$ T cell is positive for both CD3e and CD8a.

\paragraph{External cohorts.}
To test transfer across imaging platforms, we use the colorectal cancer tissue microarray of Sch\"urch \etal~\cite{schurch2020coordinated}, imaged with CODEX and H\&E-stained afterwards (35 patients).
We register the two modalities on each core, render the mIF images with the same color scheme and a calibration fitted on this cohort, and call Ki67 positivity by an Otsu threshold on the cell-level intensities.
To test transfer to a new cancer type, we also use the head and neck squamous cell carcinoma tissue microarrays of Haist \etal~\cite{haist2026lymph}, imaged with CODEX and H\&E (primary tumor regions, 76 patients), with the same tasks and rendering and the epithelium taken from the cytokeratin channel.

\paragraph{Ground truth by example.}
\Cref{fig:gt} shows the truth for one held-out field per task.
The epithelium is the Pan-CK mask (orange outline), which the model never sees; empty space (dark grey) is taken from H\&E, and stroma is the remaining tissue.
A CD8$^+$ T cell is CD3e$^+$ and CD8a$^+$, shown yellow in mIF.
The quantities stated by \hermes lie close to the truth in each example, \eg 12 stromal CD8$^+$ cells in 29\% stroma (truth: 12 and 29\%), giving 3{,}307 against 3{,}322 cells/mm$^2$.

\subsection{Tasks: three kinds of reasoning, four tasks}
\label{sec:tasks}
Each task mirrors a readout in clinical use (supplementary \cref{tab:tasks}).
We write $F$ for the field ($|F| = 0.0125$\,mm$^2$), $E$ for the epithelium mask and $S$ for the stroma, with area shares $f_E = |E|/|F|$ and $f_S = |S|/|F|$.
Cell $i$ has nucleus position $p_i$ and is epithelial if $e_i = \mathbf{1}[p_i \in E]$; $c_i$ and $k_i$ indicate a CD8$^+$ T cell and a Ki67$^+$ nucleus.

\textbf{Density} asks for the number of CD8$^+$ T cells per mm$^2$ of a compartment $C \in \{E, S\}$, a compartment-level counterpart of the densities behind the Immunoscore~\cite{pages2018immunoscore}:
\begin{equation}
y_\text{dens} = \frac{\sum_i c_i\, \mathbf{1}[p_i \in C]}{f_C\, |F|} .
\end{equation}
The area excludes empty space, and the asked compartment covers at least 15\% of the field.

\textbf{Ki67 fraction} asks for the share of epithelial nuclei that are Ki67$^+$, the Ki67 index of the International Ki67 Working Group~\cite{nielsen2021ki67}, which counts tumour nuclei only:
\begin{equation}
y_\text{frac} = 100 \cdot \frac{\sum_i k_i\, e_i}{\sum_i e_i}\ \%,
\end{equation}
on fields with at least 10 epithelial nuclei.

\textbf{Ki67 source} asks whether the Ki67$^+$ cells are mainly epithelial, the attribution a scorer must make before computing the Ki67 index, since only tumor nuclei count.
With $n_E = \sum_i k_i e_i$ and $n_O = \sum_i k_i (1 - e_i)$,
\begin{equation}
y_\text{src} = \begin{cases} \text{epithelial} & n_E > n_O,\\ \text{non-epithelial} & \text{otherwise,} \end{cases}
\end{equation}
on fields with $n_E + n_O \ge 5$ and a clear majority, $\max(n_E, n_O) \ge 0.7\,(n_E + n_O)$.
Among non-epithelial fields we take first those that still contain at least 10 epithelial nuclei.

\textbf{Regions} asks for boxes around the intra-epithelial CD8$^+$ T cells $P = \{p_i : c_i e_i = 1\}$, or ``none'' if $P$ is empty.
The true regions are the connected components $G_1, \dots, G_m$ of the graph on $P$ that links two cells closer than 20\,$\mu$m (single linkage), each boxed with a margin $\delta = 5\,\mu$m:
\begin{equation}
b_j = \bigl[\min_{p \in G_j} u - \delta,\ \min_{p \in G_j} v - \delta,\ \max_{p \in G_j} u + \delta,\ \max_{p \in G_j} v + \delta\bigr],
\end{equation}
where $p = (u, v)$ and coordinates are on a 0--1000 image scale.

Quantitative tasks therefore count marker-positive cells within an H\&E-defined compartment and normalise them by area or by the number of epithelial nuclei, localization finds where intra-epithelial marker-positive cells cluster, and attribution decides which compartment the marker-positive cells mainly belong to.
The data contain 8{,}000 questions for supervised fine-tuning, 6{,}400 for reinforcement learning, 5{,}302 for validation and 5{,}774 for testing; model comparisons use a seeded subset of 250 test questions per task.

\subsection{Guideline-structured reasoning}
\label{sec:template}
Rather than letting the model reason freely, we give it a short reasoning template that adopts three rules from clinical scoring.
First, a compartment's area is its denominator, following TIL scoring~\cite{salgado2015tils}, which reports TILs as a share of stromal area and leaves out necrosis and artefacts: epithelium and stroma are delineated on H\&E and empty space is left out.
Second, CD8$^+$ T cells are reported as a density, the quantity the Immunoscore is built on~\cite{pages2018immunoscore}, here per mm$^2$ of the compartment.
Third, the Ki67 index follows the International Ki67 in Breast Cancer Working Group~\cite{nielsen2021ki67}: staining intensity is not interpreted, and the index is the share of positive cells among invasive cancer cells.
Every step names the image it reads from and states one quantity (\cref{fig:gt}, bottom).
For Density, the model reads the epithelial, stromal and empty shares of the field from H\&E, counts the CD3$^+$CD8$^-$ cells it excludes and the CD8$^+$ cells in the asked compartment from mIF, and divides the count by the compartment area.
For Regions, it states the epithelial share, the excluded CD3$^+$CD8$^-$ cells, and the CD8$^+$ cells inside and outside the epithelium with their positions, and then groups the intra-epithelial cells into regions.
For Ki67 fraction and Ki67 source, it states the epithelial share and the number of epithelial nuclei from H\&E and the Ki67$^+$ nuclei inside and outside the epithelium from mIF; the fraction is the ratio of two of these counts, and the source follows from comparing the last two, not from the area.
Each task thus has four checkable intermediate quantities.
The template makes the reasoning short and makes every step comparable with the truth.

\paragraph{Supervised fine-tuning.}
We fill the template with the true quantities of each training field to obtain reasoning traces whose every number is correct, and fine-tune Qwen3.5-4B on them with LoRA~\cite{hu2022lora} (rank 64 on all linear layers of the language model; vision encoder frozen) for two epochs with a learning rate of $10^{-4}$, which teaches the format and the order of the steps and how to read the quantities from the two images.

\subsection{RL with a verifiable process reward}
\label{sec:grpo}
Starting from the fine-tuned model, we apply GRPO~\cite{shao2024deepseekmath}.
For each question, the policy samples a group of $G$ completions, each completion receives a scalar reward $r_i$, and its advantage is the reward normalised within the group, $\hat A_i = (r_i - \operatorname{mean}_j r_j)/\operatorname{std}_j r_j$.
A reward term affects training only when it varies across the completions of a question.
The reward has three terms,
\begin{equation}
r = r_\text{fmt} + r_\text{ans} + \lambda\, r_\text{proc}, \qquad \lambda = 0.5 .
\label{eq:reward}
\end{equation}
The format reward $r_\text{fmt}\in\{0,1\}$ checks that the completion is a reasoning block followed by one answer.

\paragraph{Answer reward.}
$r_\text{ans}\in[0,1]$ is task-specific and graded, so that closer answers earn more.
For Density, $r_\text{ans} = \max\bigl(0, 1 - |\ln\frac{\hat y + c}{y + c}| / \ln 2\bigr)$ with $c = 80$ cells/mm$^2$ (one cell in a fully tissue-covered field), which is zero once the answer is off by a factor of two.
For Ki67 fraction, $r_\text{ans} = \max(0, 1 - |\hat y - y|/20)$ with $y$ in percentage points.
For Ki67 source, $r_\text{ans}$ is 1 for the correct label and 0 otherwise.
For Regions, a ``none'' answer scores 1 only if the truth is ``none''; otherwise the score is the F1 of cell recall (share of intra-epithelial CD8$^+$ cells inside a predicted box) and box precision (share of predicted boxes that contain at least one), multiplied by $1 - 0.5\,s$, where $s$ is the share of avoidable stromal CD8$^+$ cells covered by the boxes, and by $\min(1, 2A/\hat A)$, where $A$ and $\hat A$ are the areas covered by the true and predicted boxes, so that oversized boxes are penalised.

\paragraph{Process reward.}
$r_\text{proc}$ checks the four intermediate quantities $\hat q_k$ stated in the reasoning against their truth $q_k$:
\begin{equation}
r_\text{proc} = \frac{1}{4}\sum_{k=1}^{4} \max\Bigl(0,\, 1 - \frac{|\hat q_k - q_k|}{w_k}\Bigr),
\label{eq:proc}
\end{equation}
where $w_k = 30$ percentage points for area shares and $w_k = \tau(q_k)$ for cell counts, and a quantity that is not stated scores 0.
Because the truth exists for every quantity, the process reward needs no learned reward model.

\paragraph{Tolerance curriculum.}
A strict tolerance gives almost no reward early in training, when counts are still far off, while a loose one stops rewarding improvement later.
We therefore tighten the count tolerance $\tau(q)$ in three stages, each lasting a third of training: $\tau(q) = \max(5, q)$, then $\max(3, 0.5\,q)$, then $\max(2, 0.3\,q)$.

\paragraph{Implementation.}
We train for 200 steps with 32 questions per step and $G = 8$ completions per question (sampling temperature 1, at most 768 completion tokens), training a new LoRA adapter (rank 64) on the fine-tuned model with a constant learning rate of $10^{-5}$ after 10 warm-up steps.
Training runs on four A100-80GB GPUs with vLLM~\cite{kwon2023vllm} generation.
We report the final checkpoint (step 200).

\section{Experiments}
\label{sec:exp}

\subsection{Setup}
\paragraph{Data.}
We evaluate on the five held-out Orion patients and on two external cohorts, without retraining.
Models are compared on a seeded subset of 250 questions per task on Orion (1,000 questions).
The Sch\"urch cohort uses the same sampling (874 questions, as fewer fields qualify for some tasks).
On the head and neck cohort of Haist \etal~\cite{haist2026lymph} (1,000 questions), we run the same models as on Sch\"urch.

\paragraph{Baselines.}
We compare with twelve MLLMs and with a task-specific baseline trained on the same inputs.
They comprise three proprietary models (GPT-5.6 Sol, Gemini 3.7 Flash, Claude Opus 5), three general models (Gemma-4-31B, Qwen3.8-27B, Ministral-3-14B), two medical models (MedGemma-27B~\cite{sellergren2025medgemma}, HuatuoGPT-3-27B), three pathology models (Quilt-LLaVA-7B~\cite{seyfioglu2024quiltllava}, PathGen-LLaVA-13B~\cite{sun2025pathgen}, Patho-R1-7B~\cite{zhang2025pathor1}), and the unmodified Qwen3.5-4B on which \hermes is built.
All models receive the same prompts and images, with thinking enabled where supported and no fine-tuning on our data.
The one exception is Claude Opus 5, which answers an answer-only prompt because its API refused the step-by-step prompt.

\paragraph{Metrics.}
Answers are scored with the task-specific accuracy defined in \cref{tab:main}, averaged over the four tasks, and with macro-F1 on the same tasks.
An unreadable answer counts as wrong.
Field-level answers are mapped to clinical tiers and compared with the truth by quadratic weighted $\kappa$~\cite{cohen1968weighted}.
Reasoning is scored by the share of intermediate quantities that match the cell-level truth, for open-weight models, as proprietary models hide their reasoning.

\subsection{Main results}
\label{sec:exp_main}
\Cref{fig:teaser} summarises the results on the held-out Orion patients, and \cref{tab:main} in the supplement lists all models and metrics.
\hermes (RL) has the highest accuracy on every task, tied with SFT on Ki67 source, and its average of 84.5\% compares with 74.7\% for the strongest baseline, Gemini 3.7 Flash.
Macro-F1 gives the same ranking (83.7 vs.\ 75.5).
RL improves the average over SFT from 78.8\% to 84.5\%, and all five test patients improve; Qwen3.5-4B without our training reaches 45.2\%.
Two further GRPO seeds reach 83.4\% and 82.7\% (supplementary \cref{tab:seeds}).
\hermes does this with about 190 output tokens in the guideline-structured format, against 1.2k--6.5k for the three strongest open-weight baselines; reward-based scores are in the supplement.

The advantage persists on the external Sch\"urch cohort, a different platform (\cref{tab:schurch}): 76.4\% vs.\ 67.3\% for Gemini 3.7 Flash, with RL better than SFT in 26 of 35 patients on all 2,605 Sch\"urch questions; on head and neck cancer, absent from training, \hermes reaches 76.7\% against 72.5\%.

\begin{figure*}[t]
\centering
\includegraphics[width=\linewidth]{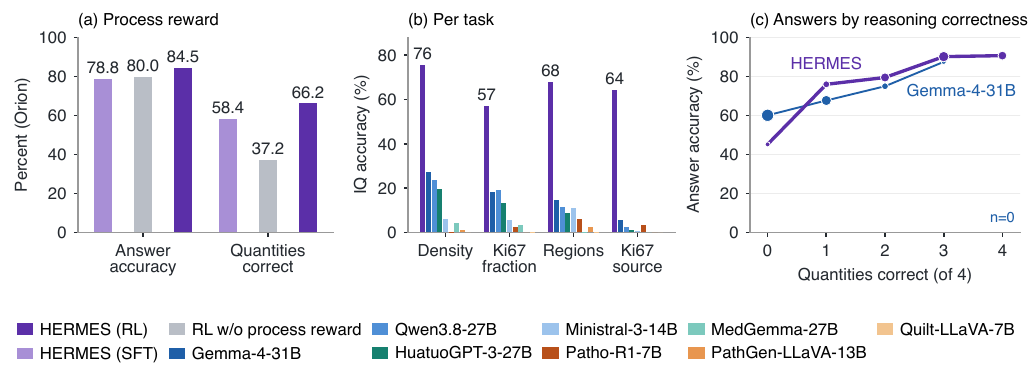}
\caption{Intermediate quantities checked against cell-level truth; baseline quantities are extracted from their reasoning by an LLM. (a) GRPO with and without the process reward, from the same SFT model (Orion; \cref{tab:ablation_process}). (b) Share of needed quantities correct per task and model (unstated counts as wrong; \cref{tab:iq}). (c) Answer accuracy vs.\ number of correct quantities per question, \hermes (RL) and Gemma-4-31B (marker area: share of questions; n=0: Gemma-4-31B never states all four correctly).}
\label{fig:iq}
\end{figure*}
\subsection{Clinical readouts}
\label{sec:exp_clinical}
Each field-level answer maps to a clinical tier, so we can ask whether the tier would be right (\cref{fig:teaser,tab:main}, quadratic weighted $\kappa$).
On Orion, \hermes leads all three readouts: 91.9 vs.\ 87.8 for Ki67 and 56.6 vs.\ 53.3 for intra-epithelial CD8 (both against Gemini 3.7 Flash), and 79.0 vs.\ 73.4 for stromal CD8 (against GPT-5.6 Sol).
On the Sch\"urch cohort, \hermes again leads on Ki67 (87.0 vs.\ 77.3) and intra-epithelial CD8 (43.9 vs.\ 32.1), and ties Gemini 3.7 Flash on stromal CD8 (58.9; \cref{tab:schurch}).
Pathology MLLMs show no meaningful agreement beyond chance ($\kappa \le 6.6$ on Orion).

\subsection{Is the reasoning correct?}
\label{sec:exp_iq}
\begin{figure}[h]
\centering
\includegraphics[width=\linewidth]{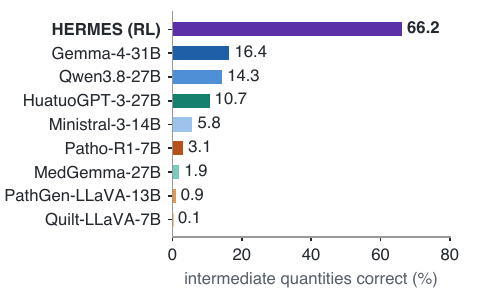}
\caption{Share of the quantities a task needs that each model states correctly (Orion; unstated counts as wrong). Baseline quantities are extracted from their reasoning by an LLM.}
\label{fig:iq_permodel}
\end{figure}

\Cref{fig:iq} summarises the reasoning check. General and medical MLLMs state only 19--36\% of the quantities that the scoring rules require, and pathology MLLMs under 10\% (supplementary \cref{fig:iq_permodel}): without guideline-structured training, they do not reason the way the readouts are scored. The strongest open-weight baseline (Gemma-4-31B) gets only 16.4\% of the required quantities right. \hermes states every quantity and 66.2\% are correct, with the largest gap on Ki67 source (64\% vs.\ 6\%).

RL improves the reasoning as well as the answer: the correct share rises from 58.4\% after SFT to 66.2\%, and 15 of the 16 task-specific quantities improve.
This depends on the process reward (\cref{fig:iq}a): GRPO from the same SFT model without it raises accuracy only from 78.8\% to 80.0\%, and the correct share falls to 37.2\%. Without it, the model also leaves the guideline steps: on the two Ki67 tasks, 96--99\% of its outputs count all nuclei on the Hoechst channel of the mIF image instead of the epithelial nuclei on H\&E, and only 1.2\% of its epithelial nucleus counts on Ki67 source are correct, against 30\% for \hermes{} (supplementary \cref{sec:suppl_ablation_process}).

Correct reasoning goes with correct answers: accuracy of \hermes rises from 45.2\% on the 31 questions with no correct quantity to 90.7\% on the 236 with all four correct (\cref{fig:iq}c).
When we edit the decisive quantity in the reasoning (\eg tripling the CD8$^+$ count or swapping the two Ki67$^+$ counts) and let the model continue, the answer changes as the edit implies in all 885 test questions, and without an edit it is reproduced in all of them: the answer is computed from the stated quantities.
Editing the image has the same effect: removing three CD8$^+$ cells from the mIF image lowers the stated count in 89.7\% of fields, against 69.6\% and 54.4\% for Gemma-4-31B and Qwen3.8-27B, while a sham edit away from T cells leaves it unchanged in 73.3\% (5.1\% and 16.4\%; supplementary \cref{sec:suppl_cf}).

\subsection{Localisation: knowing when not to box}
\label{sec:exp_regions}
Regions requires deciding whether a field contains intra-epithelial CD8$^+$ cells (122 of the 250 Orion test fields do) and, if so, boxing them.
The strongest baselines tend to draw boxes even when there is nothing to box, answering ``none'' in only 34--42\% of fields without infiltrate, whereas \hermes does so in 76.6\% while still boxing 86.1\% of fields with infiltrate; its boxes are also more precise (67.5\% vs.\ 55.9--62.4\% contain an intra-epithelial CD8$^+$ cell), and its box-level F1 is 60.3 vs.\ 48.9 at IoU $\geq 0.1$ and 46.0 vs.\ 38.3 at IoU $\geq 0.3$ against the strongest baselines (supplementary \cref{tab:regionbox}).
RL produces this balance: false ``none'' answers drop from 40.2\% after SFT to 13.9\%.

\subsection{Ablation: \hermes uses both modalities}
\label{sec:ablation}
\label{sec:exp_modality}
Masking either image while keeping the other intact lowers accuracy on every task (validation set; supplementary \cref{fig:modality}): the average falls from 86.7\% to 76.4\% without H\&E and to 50.0\% without mIF.
H\&E supplies the compartment and the denominator of the Ki67 index, mIF the identity of each cell, so \hermes needs both.

\section{Conclusion}
\label{sec:conclusion}

We let an MLLM reason jointly over H\&E and mIF in guideline steps, each grounded in one modality, and supervise the reasoning itself by checking every stated quantity against cell-level truth from the same section.
\hermes achieves the highest accuracy among twelve MLLMs on colorectal cancer and keeps it without retraining on another imaging platform and on head and neck cancer.
Its reasoning can be verified: most of its intermediate quantities are correct, its answers are computed from them, and they change as the cells change when the mIF image is edited.
The process reward is central to this: without it, GRPO barely improves accuracy over SFT and the stated quantities become less correct.
Same-section imaging thus turns pathology reasoning into a target that can be supervised step by step, and extending it to further readouts and cancer types is a natural next step.

\section*{LLM Usage Statement}
Large language models are part of the study itself: the general, medical and pathology MLLMs in \cref{sec:exp} are evaluated as baselines, and Qwen3.6-27B is used, without access to the ground truth, to extract the intermediate quantities stated in each model's reasoning (\cref{sec:exp_iq}).
Claude was used to edit and polish author-produced text and to write and debug code; the authors checked all text, code and results and take full responsibility for them.

{
    \small
    \bibliographystyle{ieeenat_fullname}
    \bibliography{main}

@String(CVPR= {IEEE Conf. Comput. Vis. Pattern Recog.})

@String(ICCV= {Int. Conf. Comput. Vis.})

@String(ICLR = {Int. Conf. Learn. Represent.})

@String(AAAI = {AAAI})

@String(CVPR  = {CVPR})

@String(ICCV  = {ICCV})

@String(ICLR  = {ICLR})

@article{lin2023orion,
  title   = {High-plex immunofluorescence imaging and traditional histology of the same tissue section for discovering image-based biomarkers},
  author  = {Lin, Jia-Ren and Chen, Yu-An and Campton, Daniel and others},
  journal = {Nature Cancer},
  volume  = {4},
  pages   = {1036--1052},
  year    = {2023}
}

@article{schurch2020coordinated,
  title   = {Coordinated cellular neighborhoods orchestrate antitumoral immunity at the colorectal cancer invasive front},
  author  = {Sch{\"u}rch, Christian M. and Bhate, Salil S. and Barlow, Graham L. and others},
  journal = {Cell},
  volume  = {182},
  number  = {5},
  pages   = {1341--1359},
  year    = {2020}
}

@article{pages2018immunoscore,
  title   = {International validation of the consensus {Immunoscore} for the classification of colon cancer: a prognostic and accuracy study},
  author  = {Pag{\`e}s, Franck and Mlecnik, Bernhard and Marliot, Florence and others},
  journal = {The Lancet},
  volume  = {391},
  number  = {10135},
  pages   = {2128--2139},
  year    = {2018}
}

@article{haist2026lymph,
  title   = {Lymph node colonization induces tissue remodeling via immunosuppressive fibroblast-myeloid cell niches supporting metastatic tolerance},
  author  = {Haist, Maximilian and others},
  journal = {Cancer Cell},
  volume  = {44},
  number  = {3},
  pages   = {604--623},
  doi     = {10.1016/j.ccell.2026.01.003},
  year    = {2026}
}

@article{salgado2015tils,
  title   = {The evaluation of tumor-infiltrating lymphocytes ({TILs}) in breast cancer: recommendations by an {International TILs Working Group} 2014},
  author  = {Salgado, Roberto and Denkert, Carsten and Demaria, Sandra and others},
  journal = {Annals of Oncology},
  volume  = {26},
  number  = {2},
  pages   = {259--271},
  year    = {2015}
}

@article{nielsen2021ki67,
  title   = {Assessment of {Ki67} in breast cancer: updated recommendations from the {International Ki67 in Breast Cancer Working Group}},
  author  = {Nielsen, Torsten O. and Leung, Samuel C. Y. and Rimm, David L. and others},
  journal = {Journal of the National Cancer Institute},
  volume  = {113},
  number  = {7},
  pages   = {808--819},
  year    = {2021}
}

@inproceedings{zhang2025pathor1,
  title     = {{Patho-R1}: A multimodal reinforcement learning-based pathology expert reasoner},
  author    = {Zhang, Wenchuan and others},
  booktitle = {AAAI},
  volume    = {40},
  number    = {33},
  pages     = {28418--28426},
  doi       = {10.1609/aaai.v40i33.40071},
  year      = {2026}
}

@article{guo2025deepseekr1,
  title   = {{DeepSeek-R1} incentivizes reasoning in {LLMs} through reinforcement learning},
  author  = {Guo, Daya and Yang, Dejian and Zhang, Haowei and others},
  journal = {Nature},
  volume  = {645},
  pages   = {633--638},
  year    = {2025}
}

@article{shao2024deepseekmath,
  title   = {{DeepSeekMath}: Pushing the limits of mathematical reasoning in open language models},
  author  = {Shao, Zhihong and Wang, Peiyi and Zhu, Qihao and others},
  journal = {arXiv preprint arXiv:2402.03300},
  year    = {2024}
}

@inproceedings{seyfioglu2024quiltllava,
  title     = {{Quilt-LLaVA}: Visual instruction tuning by extracting localized narratives from open-source histopathology videos},
  author    = {Seyfioglu, Mehmet Saygin and Ikezogwo, Wisdom O. and Ghezloo, Fatemeh and Krishna, Ranjay and Shapiro, Linda},
  booktitle = {CVPR},
  year      = {2024}
}

@inproceedings{sun2025pathgen,
  title     = {{PathGen-1.6M}: 1.6 million pathology image-text pairs generation through multi-agent collaboration},
  author    = {Sun, Yuxuan and others},
  booktitle = {ICLR},
  year      = {2025}
}

@article{sellergren2025medgemma,
  title   = {{MedGemma} technical report},
  author  = {Sellergren, Andrew and others},
  journal = {arXiv preprint arXiv:2507.05201},
  year    = {2025}
}

@article{lu2024pathchat,
  title   = {A multimodal generative {AI} copilot for human pathology},
  author  = {Lu, Ming Y. and Chen, Bowen and Williamson, Drew F. K. and others},
  journal = {Nature},
  volume  = {634},
  pages   = {466--473},
  year    = {2024}
}

@inproceedings{li2023llavamed,
  title     = {{LLaVA-Med}: Training a large language-and-vision assistant for biomedicine in one day},
  author    = {Li, Chunyuan and Wong, Cliff and Zhang, Sheng and others},
  booktitle = {NeurIPS},
  year      = {2023}
}

@article{lu2024conch,
  title   = {A visual-language foundation model for computational pathology},
  author  = {Lu, Ming Y. and Chen, Bowen and Williamson, Drew F. K. and others},
  journal = {Nature Medicine},
  volume  = {30},
  pages   = {863--874},
  year    = {2024}
}

@article{huang2023plip,
  title   = {A visual--language foundation model for pathology image analysis using medical {Twitter}},
  author  = {Huang, Zhi and Bianchi, Federico and Yuksekgonul, Mert and Montine, Thomas J. and Zou, James},
  journal = {Nature Medicine},
  volume  = {29},
  pages   = {2307--2316},
  year    = {2023}
}

@inproceedings{ikezogwo2023quilt,
  title     = {{Quilt-1M}: One million image-text pairs for histopathology},
  author    = {Ikezogwo, Wisdom O. and Seyfioglu, Mehmet Saygin and Ghezloo, Fatemeh and others},
  booktitle = {NeurIPS Datasets and Benchmarks},
  year      = {2023}
}

@article{chen2024uni,
  title   = {Towards a general-purpose foundation model for computational pathology},
  author  = {Chen, Richard J. and Ding, Tong and Lu, Ming Y. and others},
  journal = {Nature Medicine},
  volume  = {30},
  pages   = {850--862},
  year    = {2024}
}

@article{xu2024gigapath,
  title   = {A whole-slide foundation model for digital pathology from real-world data},
  author  = {Xu, Hanwen and Usuyama, Naoto and Bagga, Jaspreet and others},
  journal = {Nature},
  volume  = {630},
  pages   = {181--188},
  year    = {2024}
}

@inproceedings{wei2022cot,
  title     = {Chain-of-thought prompting elicits reasoning in large language models},
  author    = {Wei, Jason and Wang, Xuezhi and Schuurmans, Dale and others},
  booktitle = {NeurIPS},
  year      = {2022}
}

@article{lai2025medr1,
  title   = {{Med-R1}: Reinforcement learning for generalizable medical reasoning in vision-language models},
  author  = {Lai, Yuxiang and others},
  journal = {arXiv preprint arXiv:2503.13939},
  year    = {2025}
}

@inproceedings{pan2025medvlmr1,
  title   = {{MedVLM-R1}: Incentivizing medical reasoning capability of vision-language models via reinforcement learning},
  author  = {Pan, Jiazhen and others},
  booktitle = {Medical Image Computing and Computer Assisted Intervention (MICCAI)},
  year    = {2025}
}

@inproceedings{huang2025visionr1,
  title     = {{Vision-R1}: Incentivizing reasoning capability in multimodal large language models},
  author    = {Huang, Wenxuan and Jia, Bohan and Zhai, Zijie and Cao, Shaosheng and Ye, Zheyu and Zhao, Fei and Xu, Zhe and Tang, Xu and Hu, Yao and Lin, Shaohui},
  booktitle = {ICLR},
  year      = {2026}
}

@inproceedings{liu2025visualrft,
  title   = {{Visual-RFT}: Visual reinforcement fine-tuning},
  author  = {Liu, Ziyu and others},
  booktitle = {ICCV},
  year    = {2025}
}

@article{uesato2022process,
  title   = {Solving math word problems with process- and outcome-based feedback},
  author  = {Uesato, Jonathan and Kushman, Nate and Kumar, Ramana and others},
  journal = {arXiv preprint arXiv:2211.14275},
  year    = {2022}
}

@inproceedings{lightman2024verify,
  title     = {Let's verify step by step},
  author    = {Lightman, Hunter and Kosaraju, Vineet and Burda, Yura and others},
  booktitle = {ICLR},
  year      = {2024}
}

@inproceedings{turpin2023unfaithful,
  title     = {Language models don't always say what they think: Unfaithful explanations in chain-of-thought prompting},
  author    = {Turpin, Miles and Michael, Julian and Perez, Ethan and Bowman, Samuel R.},
  booktitle = {NeurIPS},
  year      = {2023}
}

@article{lanham2023faithfulness,
  title   = {Measuring faithfulness in chain-of-thought reasoning},
  author  = {Lanham, Tamera and Chen, Anna and Radhakrishnan, Ansh and others},
  journal = {arXiv preprint arXiv:2307.13702},
  year    = {2023}
}

@article{goltsev2018codex,
  title   = {Deep profiling of mouse splenic architecture with {CODEX} multiplexed imaging},
  author  = {Goltsev, Yury and Samusik, Nikolay and Kennedy-Darling, Julia and others},
  journal = {Cell},
  volume  = {174},
  number  = {4},
  pages   = {968--981},
  year    = {2018}
}

@article{lin2018cycif,
  title   = {Highly multiplexed immunofluorescence imaging of human tissues and tumors using t-{CyCIF} and conventional optical microscopes},
  author  = {Lin, Jia-Ren and Izar, Benjamin and Wang, Shu and others},
  journal = {eLife},
  volume  = {7},
  pages   = {e31657},
  year    = {2018}
}

@article{rivenson2019virtual,
  title   = {Virtual histological staining of unlabelled tissue-autofluorescence images via deep learning},
  author  = {Rivenson, Yair and Wang, Hongda and Wei, Zhensong and others},
  journal = {Nature Biomedical Engineering},
  volume  = {3},
  pages   = {466--477},
  year    = {2019}
}

@article{saltz2018tils,
  title   = {Spatial organization and molecular correlation of tumor-infiltrating lymphocytes using deep learning on pathology images},
  author  = {Saltz, Joel and Gupta, Rajarsi and Hou, Le and others},
  journal = {Cell Reports},
  volume  = {23},
  number  = {1},
  pages   = {181--193},
  year    = {2018}
}

@article{valanarasu2025gigatime,
  title   = {Multimodal {AI} generates virtual population for tumor microenvironment modeling},
  author  = {Valanarasu, Jeya Maria Jose and others},
  journal = {Cell},
  volume  = {189},
  number  = {2},
  pages   = {386--400},
  doi     = {10.1016/j.cell.2025.11.016},
  year    = {2026}
}

@article{balezo2025miphei,
  title   = {{MIPHEI-ViT}: Multiplex immunofluorescence prediction from {H\&E} images using {ViT} foundation models},
  author  = {Balezo, Guillaume and Trullo, Roger and Pla Planas, Albert and Decenci{\`e}re, Etienne and Walter, Thomas},
  journal = {arXiv preprint arXiv:2505.10294},
  year    = {2025}
}

@article{binnewies2018time,
  title={Understanding the tumor immune microenvironment ({TIME}) for effective therapy},
  author={Binnewies, Mikhail and Roberts, Edward W. and Kersten, Kelly and Chan, Vincent and Fearon, Douglas F. and Merad, Miriam and Coussens, Lisa M. and Gabrilovich, Dmitry I. and Ostrand-Rosenberg, Suzanne and Hedrick, Catherine C. and others},
  journal={Nature Medicine},
  volume={24},
  number={5},
  pages={541--550},
  year={2018}
}

@article{chen2017elements,
  title={Elements of cancer immunity and the cancer--immune set point},
  author={Chen, Daniel S. and Mellman, Ira},
  journal={Nature},
  volume={541},
  number={7637},
  pages={321--330},
  year={2017}
}

@article{galon2006type,
  title   = {Type, density, and location of immune cells within human colorectal tumors predict clinical outcome},
  author  = {Galon, J{\'e}r{\^o}me and Costes, Anne and Sanchez-Cabo, Fatima and others},
  journal = {Science},
  volume  = {313},
  number  = {5795},
  pages   = {1960--1964},
  year    = {2006}
}

@article{fridman2012immune,
  title   = {The immune contexture in human tumours: impact on clinical outcome},
  author  = {Fridman, Wolf Herman and Pag{\`e}s, Franck and Saut{\`e}s-Fridman, Catherine and Galon, J{\'e}r{\^o}me},
  journal = {Nature Reviews Cancer},
  volume  = {12},
  number  = {4},
  pages   = {298--306},
  year    = {2012}
}

@article{hendry2017tils,
  title   = {Assessing tumor-infiltrating lymphocytes in solid tumors: a practical review for pathologists and proposal for a standardized method from the {International Immunooncology Biomarkers Working Group}: Part 1},
  author  = {Hendry, Shona and Salgado, Roberto and Gevaert, Thomas and others},
  journal = {Advances in Anatomic Pathology},
  volume  = {24},
  number  = {5},
  pages   = {235--251},
  year    = {2017}
}

@article{dowsett2011ki67,
  title   = {Assessment of {Ki67} in breast cancer: recommendations from the {International Ki67 in Breast Cancer Working Group}},
  author  = {Dowsett, Mitch and Nielsen, Torsten O. and A'Hern, Roger and others},
  journal = {Journal of the National Cancer Institute},
  volume  = {103},
  number  = {22},
  pages   = {1656--1664},
  year    = {2011}
}

@article{lewis2021spatial,
  title   = {Spatial omics and multiplexed imaging to explore cancer biology},
  author  = {Lewis, Sabrina M. and Asselin-Labat, Marie-Liesse and Nguyen, Quan and others},
  journal = {Nature Methods},
  volume  = {18},
  number  = {9},
  pages   = {997--1012},
  year    = {2021}
}

@article{giesen2014imc,
  title   = {Highly multiplexed imaging of tumor tissues with subcellular resolution by mass cytometry},
  author  = {Giesen, Charlotte and Wang, Hao A. O. and Schapiro, Denis and others},
  journal = {Nature Methods},
  volume  = {11},
  number  = {4},
  pages   = {417--422},
  year    = {2014}
}

@article{keren2018mibi,
  title   = {A structured tumor-immune microenvironment in triple negative breast cancer revealed by multiplexed ion beam imaging},
  author  = {Keren, Leeat and Bosse, Marc and Marquez, Diana and others},
  journal = {Cell},
  volume  = {174},
  number  = {6},
  pages   = {1373--1387},
  year    = {2018}
}

@article{greenwald2022mesmer,
  title   = {Whole-cell segmentation of tissue images with human-level performance using large-scale data annotation and deep learning},
  author  = {Greenwald, Noah F. and Miller, Geneva and Moen, Erick and others},
  journal = {Nature Biotechnology},
  volume  = {40},
  number  = {4},
  pages   = {555--565},
  year    = {2022}
}

@article{graham2019hovernet,
  title   = {{HoVer-Net}: Simultaneous segmentation and classification of nuclei in multi-tissue histology images},
  author  = {Graham, Simon and Vu, Quoc Dang and Raza, Shan E. Ahmed and others},
  journal = {Medical Image Analysis},
  volume  = {58},
  pages   = {101563},
  year    = {2019}
}

@article{campanella2019clinical,
  title   = {Clinical-grade computational pathology using weakly supervised deep learning on whole slide images},
  author  = {Campanella, Gabriele and Hanna, Matthew G. and Geneslaw, Luke and others},
  journal = {Nature Medicine},
  volume  = {25},
  number  = {8},
  pages   = {1301--1309},
  year    = {2019}
}

@article{kather2019msi,
  title   = {Deep learning can predict microsatellite instability directly from histology in gastrointestinal cancer},
  author  = {Kather, Jakob Nikolas and Pearson, Alexander T. and Halama, Niels and others},
  journal = {Nature Medicine},
  volume  = {25},
  number  = {7},
  pages   = {1054--1056},
  year    = {2019}
}

@article{vorontsov2024virchow,
  title   = {A foundation model for clinical-grade computational pathology and rare cancers detection},
  author  = {Vorontsov, Eugene and Bozkurt, Alican and Casson, Adam and others},
  journal = {Nature Medicine},
  volume  = {30},
  number  = {10},
  pages   = {2924--2935},
  year    = {2024}
}

@inproceedings{liu2023llava,
  title     = {Visual instruction tuning},
  author    = {Liu, Haotian and Li, Chunyuan and Wu, Qingyang and Lee, Yong Jae},
  booktitle = {NeurIPS},
  year      = {2023}
}

@article{tu2024medpalmm,
  title   = {Towards generalist biomedical {AI}},
  author  = {Tu, Tao and Azizi, Shekoofeh and Driess, Danny and others},
  journal = {NEJM AI},
  volume  = {1},
  number  = {3},
  year    = {2024}
}

@inproceedings{chen2024huatuogptvision,
  title     = {{HuatuoGPT-Vision}, towards injecting medical visual knowledge into multimodal {LLMs} at scale},
  author    = {Chen, Junying and Gui, Chi and Ouyang, Ruyi and others},
  booktitle = {EMNLP},
  year      = {2024}
}

@inproceedings{sun2024pathasst,
  title     = {{PathAsst}: A generative foundation {AI} assistant towards artificial general intelligence of pathology},
  author    = {Sun, Yuxuan and Zhu, Chenglu and Zheng, Sunyi and others},
  booktitle = {AAAI},
  year      = {2024}
}

@inproceedings{chen2025slidechat,
  title     = {{SlideChat}: A large vision-language assistant for whole-slide pathology image understanding},
  author    = {Chen, Ying and Wang, Guoan and Ji, Yuanfeng and others},
  booktitle = {CVPR},
  year      = {2025}
}

@inproceedings{ouyang2022instructgpt,
  title     = {Training language models to follow instructions with human feedback},
  author    = {Ouyang, Long and Wu, Jeffrey and Jiang, Xu and others},
  booktitle = {NeurIPS},
  year      = {2022}
}

@inproceedings{wang2024mathshepherd,
  title     = {{Math-Shepherd}: Verify and reinforce {LLMs} step-by-step without human annotations},
  author    = {Wang, Peiyi and Li, Lei and Shao, Zhihong and others},
  booktitle = {ACL},
  year      = {2024}
}

@inproceedings{hu2022lora,
  title     = {{LoRA}: Low-rank adaptation of large language models},
  author    = {Hu, Edward J. and Shen, Yelong and Wallis, Phillip and others},
  booktitle = {ICLR},
  year      = {2022}
}

@inproceedings{kwon2023vllm,
  title     = {Efficient memory management for large language model serving with {PagedAttention}},
  author    = {Kwon, Woosuk and Li, Zhuohan and Zhuang, Siyuan and others},
  booktitle = {SOSP},
  year      = {2023}
}

@inproceedings{jacovi2020faithful,
  title     = {Towards faithfully interpretable {NLP} systems: How should we define and evaluate faithfulness?},
  author    = {Jacovi, Alon and Goldberg, Yoav},
  booktitle = {ACL},
  year      = {2020}
}

@inproceedings{lyu2023faithfulcot,
  title     = {Faithful chain-of-thought reasoning},
  author    = {Lyu, Qing and Havaldar, Shreya and Stein, Adam and others},
  booktitle = {IJCNLP-AACL},
  year      = {2023}
}

@article{rudin2019stop,
  title   = {Stop explaining black box machine learning models for high stakes decisions and use interpretable models instead},
  author  = {Rudin, Cynthia},
  journal = {Nature Machine Intelligence},
  volume  = {1},
  number  = {5},
  pages   = {206--215},
  year    = {2019}
}

@article{cohen1968weighted,
  title   = {Weighted kappa: Nominal scale agreement provision for scaled disagreement or partial credit},
  author  = {Cohen, Jacob},
  journal = {Psychological Bulletin},
  volume  = {70},
  number  = {4},
  pages   = {213--220},
  year    = {1968}
}

@article{shaban2025foundation,
  title   = {A foundation model for spatial proteomics},
  author  = {Shaban, Muhammad and Chang, Yuzhou and Qiu, Huaying and Yeo, Yao Yu and Song, Andrew H. and Jaume, Guillaume and Wang, Yuchen and Weishaupt, Luca L. and Ding, Tong and Vaidya, Anurag and others},
  journal = {arXiv preprint arXiv:2506.03373},
  year    = {2025}
}
}

\clearpage
\setcounter{page}{1}
\setcounter{section}{0}\renewcommand{\thesection}{\Alph{section}}
\twocolumn[{%
\centering
{\Large\textbf{\thetitle}\\[0.5em]Supplementary Material\\[3em]}
\begin{minipage}{0.78\textwidth}
{\large\textbf{Contents}}\par\vspace{1em}
\setlength{\tabcolsep}{6pt}\renewcommand{\arraystretch}{1.6}
\begin{tabularx}{\linewidth}{@{}l X r@{}}
\ref{sec:suppl_main}. & \hyperref[sec:suppl_main]{Full results on Orion} & \cref{tab:main} \\
\ref{sec:suppl_iq}. & \hyperref[sec:suppl_iq]{Intermediate quantities per task} & \cref{fig:iq_permodel}, \cref{tab:iq} \\
\ref{sec:suppl_tasks}. & \hyperref[sec:suppl_tasks]{Task overview} & \cref{tab:tasks} \\
\ref{sec:suppl_modality}. & \hyperref[sec:suppl_modality]{Modality masking} & \cref{fig:modality} \\
\ref{sec:suppl_external_examples}. & \hyperref[sec:suppl_external_examples]{Examples on the external cohorts} & Figs.~\ref{fig:schurch_ex1}--\ref{fig:haist_ex3} \\
\ref{sec:suppl_reward}. & \hyperref[sec:suppl_reward]{Reward-based results} & \cref{tab:reward} \\
\ref{sec:suppl_ablation_process}. & \hyperref[sec:suppl_ablation_process]{GRPO without the process reward} & Tabs.~\ref{tab:ablation_process}, \ref{tab:ablation_ext} \\
\ref{sec:suppl_fmbase}. & \hyperref[sec:suppl_fmbase]{Task-specific foundation-model baseline} & \cref{tab:fmbase} \\
\ref{sec:suppl_patient}. & \hyperref[sec:suppl_patient]{Patient-level analysis} & Tab.~\ref{tab:patient}, Fig.~\ref{fig:patient} \\
\ref{sec:suppl_cf}. & \hyperref[sec:suppl_cf]{Visual counterfactuals} & \cref{tab:cf} \\
\ref{sec:suppl_regionbox}. & \hyperref[sec:suppl_regionbox]{Box-level results on Regions} & \cref{tab:regionbox} \\
\ref{sec:suppl_seeds}. & \hyperref[sec:suppl_seeds]{GRPO seeds} & \cref{tab:seeds} \\
\end{tabularx}
\end{minipage}
}]
\clearpage

\section{Full results on Orion}
\label{sec:suppl_main}
\Cref{tab:main} gives every model and metric behind \cref{fig:teaser}.

\begin{table*}[t]
\centering\small
\setlength{\tabcolsep}{2.8pt}
\providecommand{\hd}[1]{\begin{tabular}[b]{@{}c@{}}#1\end{tabular}}
\caption{Main results on Orion (250 questions per task; values $\times 100$ except tokens). Accuracy: density within $2\times$ of truth, regions whether the field contains intra-epithelial CD8$^+$ T cells (balanced accuracy; box quality in \cref{tab:regionbox}), Ki67 source accuracy, Ki67 fraction within 10\,pp. Macro-F1: mean over the four tasks, density and Ki67 fraction on the three clinical tiers. Clinical grades: quadratic weighted $\kappa$ against truth tiers. Proprietary models hide their reasoning, so tokens are not compared.}
\label{tab:main}
\begin{tabular}{@{}l cccccc ccc r@{}}
\toprule
 & \multicolumn{5}{c}{Answer accuracy} & & \multicolumn{3}{c}{Clinical grade agreement} & \\
\cmidrule(lr){2-6}\cmidrule(lr){8-10}
Model & Density & Regions & \hd{Ki67\\source} & \hd{Ki67\\fraction} & Average & \hd{Macro-\\F1} & Ki67 & \hd{Intra-epi.\\CD8} & \hd{Stromal\\CD8} & \hd{Output\\tokens} \\
\midrule
\rowcolor{gray!12}\multicolumn{11}{c}{\emph{Proprietary MLLMs}} \\
GPT-5.6 Sol & 64.8 & 66.8 & 73.2 & 80.8 & 71.4 & 69.6 & 85.7 & 49.2 & 73.4 & -- \\
Gemini 3.7 Flash & 68.0 & 69.9 & 85.6 & 75.2 & 74.7 & 75.5 & 87.8 & 53.3 & 71.3 & -- \\
Claude Opus 5 & 56.0 & 56.1 & 66.0 & 52.4 & 57.6 & 56.9 & 63.7 & 29.9 & 50.1 & -- \\
\rowcolor{gray!12}\multicolumn{11}{c}{\emph{General MLLMs}} \\
Gemma-4-31B & 45.2 & 67.9 & 80.4 & 68.0 & 65.4 & 67.6 & 80.0 & 27.6 & 27.1 & 3,158 \\
Qwen3.8-27B & 52.0 & 68.4 & 71.2 & 67.2 & 64.7 & 67.0 & 82.6 & 31.1 & 55.1 & 6,455 \\
Ministral-3-14B & 41.6 & 52.8 & 59.6 & 30.4 & 46.1 & 42.3 & 15.1 & 14.0 & 17.6 & 288 \\
\rowcolor{gray!12}\multicolumn{11}{c}{\emph{Medical MLLMs}} \\
MedGemma-27B & 23.2 & 50.8 & 50.4 & 19.6 & 36.0 & 30.1 & -1.8 & -2.8 & 2.6 & 1,572 \\
HuatuoGPT-3-27B & 52.8 & 68.7 & 67.6 & 66.0 & 63.8 & 65.3 & 82.6 & 26.8 & 47.4 & 1,177 \\
\rowcolor{gray!12}\multicolumn{11}{c}{\emph{Pathology MLLMs}} \\
Quilt-LLaVA-7B & 6.4 & 49.6 & 2.8 & 2.4 & 15.3 & 28.4 & 0.0 & 3.5 & -5.5 & 176 \\
PathGen-LLaVA-13B & 1.2 & 50.4 & 0.0 & 0.0 & 12.9 & 28.2 & -26.5 & -9.7 & 6.6 & 65 \\
Patho-R1-7B & 26.8 & 51.4 & 27.6 & 19.2 & 31.3 & 37.9 & -2.6 & -5.4 & -8.8 & 250 \\
\midrule
Qwen3.5-4B (base) & 28.4 & 59.2 & 60.4 & 32.8 & 45.2 & 53.5 & 65.3 & 6.4 & 10.0 & 10,354 \\
\hermes (SFT) & 65.2 & 74.5 & \textbf{94.4} & 81.2 & 78.8 & 80.0 & 90.3 & 54.1 & 73.8 & 182 \\
\hermes (RL) & \textbf{76.0} & \textbf{81.3} & \textbf{94.4} & \textbf{86.4} & \textbf{84.5} & \textbf{83.7} & \textbf{91.9} & \textbf{56.6} & \textbf{79.0} & 186 \\
\bottomrule
\end{tabular}
\end{table*}

\section{Intermediate quantities per task}
\label{sec:suppl_iq}
\begin{table}[h]
\centering\small
\setlength{\tabcolsep}{3pt}
\resizebox{\linewidth}{!}{%
\begin{tabular}{@{}lcccccc@{}}
\toprule
Model & Dens. & Reg. & Src. & Frac. & Avg & Stated \\
\midrule
\multicolumn{7}{@{}l}{\emph{Extracted from free-form reasoning by an LLM judge}} \\
Gemma-4-31B     & .273 & .146 & .057 & .181 & .164 & 35.8\% \\
Qwen3.8-27B     & .238 & .116 & .026 & .192 & .143 & 28.4\% \\
Ministral-3-14B & .060 & .109 & .005 & .056 & .058 & 24.6\% \\
HuatuoGPT-3-27B & .195 & .086 & .012 & .134 & .107 & 23.6\% \\
MedGemma-27B    & .043 & .003 & .000 & .032 & .019 & 18.7\% \\
Patho-R1-7B     & .004 & .061 & .035 & .023 & .031 & 8.9\% \\
PathGen-LLaVA-13B& .011 & .025 & .000 & .000 & .009 & 1.8\% \\
Quilt-LLaVA-7B  & .000 & .003 & .000 & .001 & .001 & 0.7\% \\
\midrule
\hermes-SFT     & .661 & .620 & .571 & .485 & .584 & 100\% \\
\hermes         & \textbf{.757} & \textbf{.678} & \textbf{.642} & \textbf{.570} & \textbf{.662} & 100\% \\
\bottomrule
\end{tabular}}
\caption{Intermediate-quantity accuracy per task (complement to \cref{fig:iq}): share of the four quantities a task needs that match the cell-level truth (counts within $\max(1, 10\%)$, area shares within 10\,pp; an unstated quantity counts as wrong). Stated: share of the needed quantities a model states at all.}
\label{tab:iq}
\end{table}

\section{Task overview}
\label{sec:suppl_tasks}
\begin{table}[h]
\centering\footnotesize
\setlength{\tabcolsep}{3pt}\renewcommand{\arraystretch}{1.2}
\begin{tabular}{@{}>{\raggedright\arraybackslash}p{1.7cm}>{\raggedright\arraybackslash}p{1.6cm}>{\raggedright\arraybackslash}p{2.7cm}>{\raggedright\arraybackslash}p{1.6cm}@{}}
\toprule
Reasoning & Task & Question & Clinical readout \\
\midrule
Quantitative (how much?) & \textbf{Density} & CD8$^+$ cells per mm$^2$ of epithelium or stroma & Immuno\-score density \\
 & \textbf{Ki67 fraction} & \% Ki67$^+$ among epithelial nuclei & Ki67 index (IKWG) \\
\addlinespace
Localization (where?) & \textbf{Regions} & boxes around intra-epithelial CD8$^+$ cells, or ``none'' & intra-epithelial TILs \\
\addlinespace
Attribution (which cells?) & \textbf{Ki67 source} & are Ki67$^+$ cells mainly epithelial or non-epithelial? & gate for the Ki67 index \\
\bottomrule
\end{tabular}
\caption{Four tasks over three kinds of reasoning: quantitative (Density, Ki67 fraction), localization (Regions) and attribution (Ki67 source). Every task needs both modalities: H\&E for the compartment, mIF for the marker.}
\label{tab:tasks}
\end{table}

\section{Modality masking}
\label{sec:suppl_modality}
\begin{figure*}[h]
  \centering
  \includegraphics[width=\linewidth]{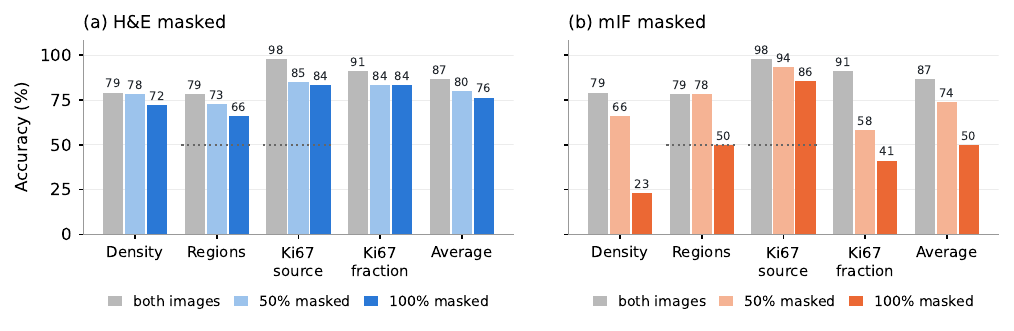}
  \caption{\hermes uses H\&E and mIF together: masking either image lowers accuracy on every task. Half or all of H\&E (a) or mIF (b) is masked while the other image stays intact; 100\% removes that image. Grey: both images; light and dark: 50\% and 100\% of the image masked; dotted: chance for the two balanced binary tasks (Regions, Ki67 source). \hermes (RL); metrics as in \cref{tab:main}.}
  \label{fig:modality}
\end{figure*}

To test whether \hermes reasons over both images, half or all of one image is masked while the other stays intact (\cref{fig:modality}; validation set, 250 questions per task).
Removing either modality lowers accuracy on every task: the average falls from 86.7\% with both images to 76.4\% without H\&E and to 50.0\% without mIF.
H\&E supplies the compartment, \ie where the epithelium ends and the stroma begins, and the denominator of the Ki67 index; mIF supplies the identity of each cell and is needed for every count.
The same holds after SFT and after RL.

\section{Examples on the external cohorts}
\label{sec:suppl_external_examples}
\begin{figure*}[p]
  \centering
  \includegraphics[width=\linewidth]{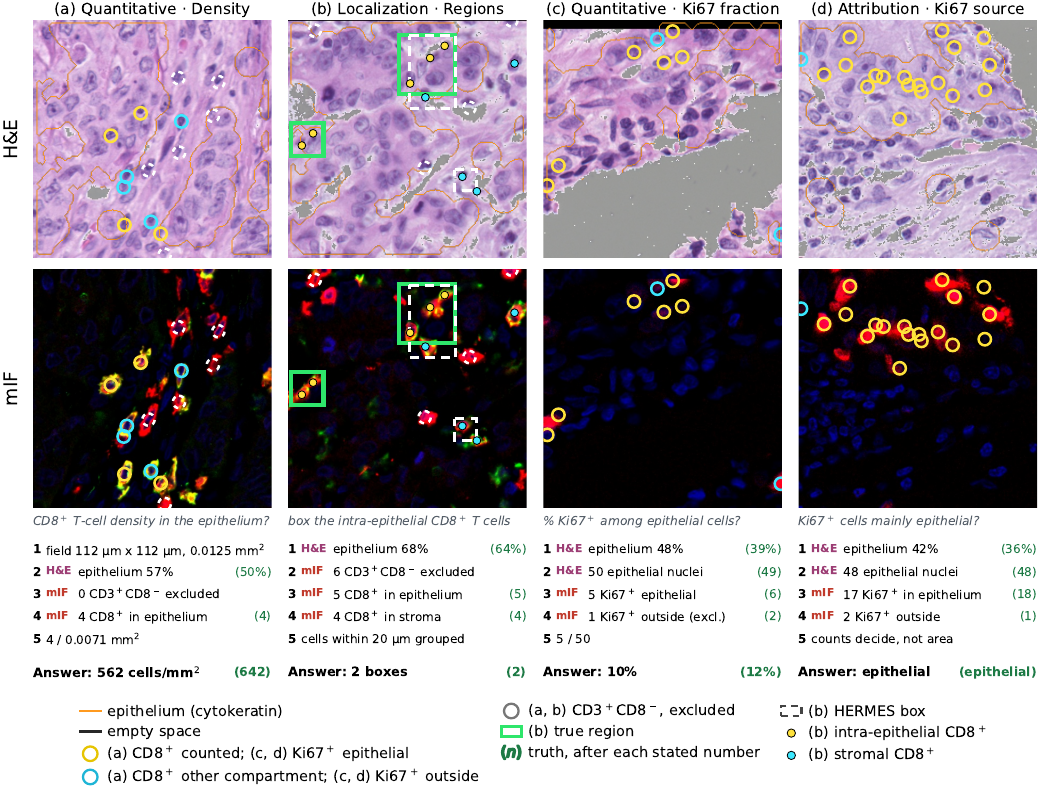}
  \caption{Sch\"urch cohort (colorectal cancer, CODEX), example set 1 of 2: one test field per task, laid out as \cref{fig:gt}. \hermes is the model trained on Orion, without retraining. Fields are chosen among correct answers whose stated counts are within one cell of the truth (Ki67 fraction within 3\,pp, density within 15\%), with 30--70\% epithelium and one patient per task and set. Steps without a bracket have no truth in this cohort.}
  \label{fig:schurch_ex1}
\end{figure*}

\begin{figure*}[p]
  \centering
  \includegraphics[width=\linewidth]{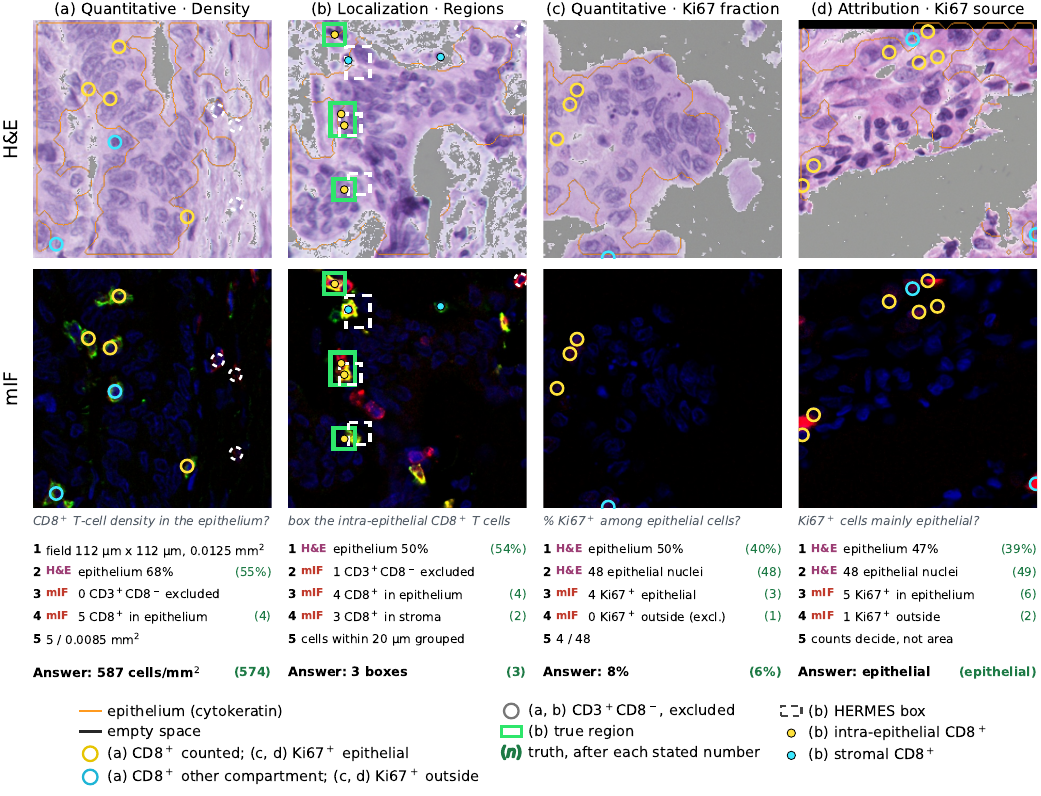}
  \caption{Sch\"urch cohort (colorectal cancer, CODEX), example set 2 of 2: one test field per task, laid out as \cref{fig:gt}. \hermes is the model trained on Orion, without retraining. Fields are chosen among correct answers whose stated counts are within one cell of the truth (Ki67 fraction within 3\,pp, density within 15\%), with 30--70\% epithelium and one patient per task and set. Steps without a bracket have no truth in this cohort.}
  \label{fig:schurch_ex2}
\end{figure*}

\begin{figure*}[p]
  \centering
  \includegraphics[width=\linewidth]{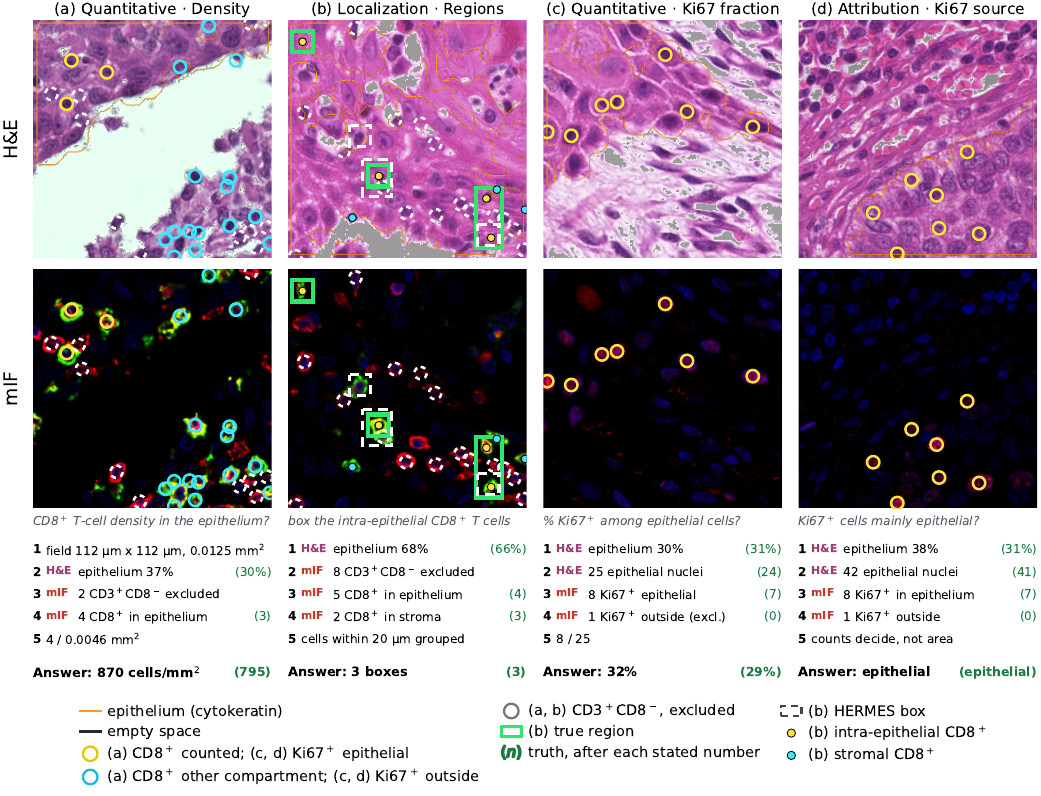}
  \caption{Haist cohort (head and neck cancer, CODEX), example set 1 of 3: one test field per task, laid out as \cref{fig:gt}. \hermes is the model trained on Orion, without retraining. Fields are chosen among correct answers whose stated counts are within one cell of the truth (Ki67 fraction within 3\,pp, density within 15\%), with 30--70\% epithelium and one patient per task and set. Steps without a bracket have no truth in this cohort.}
  \label{fig:haist_ex1}
\end{figure*}

\begin{figure*}[p]
  \centering
  \includegraphics[width=\linewidth]{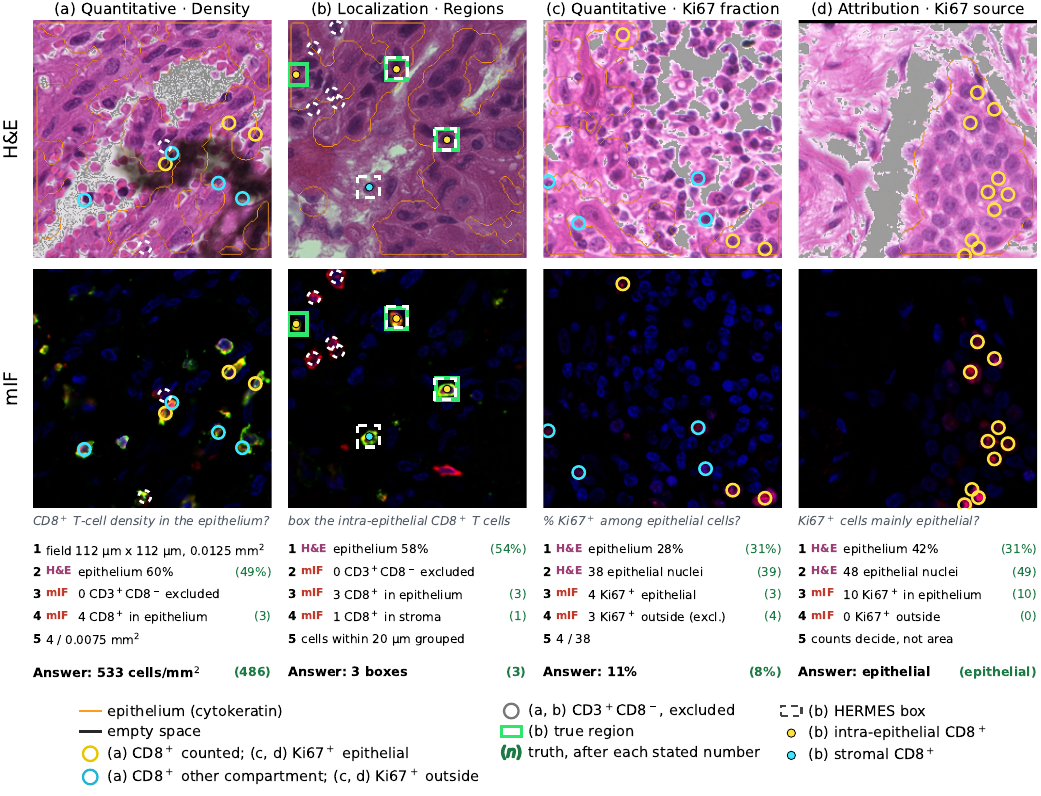}
  \caption{Haist cohort (head and neck cancer, CODEX), example set 2 of 3: one test field per task, laid out as \cref{fig:gt}. \hermes is the model trained on Orion, without retraining. Fields are chosen among correct answers whose stated counts are within one cell of the truth (Ki67 fraction within 3\,pp, density within 15\%), with 30--70\% epithelium and one patient per task and set. Steps without a bracket have no truth in this cohort.}
  \label{fig:haist_ex2}
\end{figure*}

\begin{figure*}[p]
  \centering
  \includegraphics[width=\linewidth]{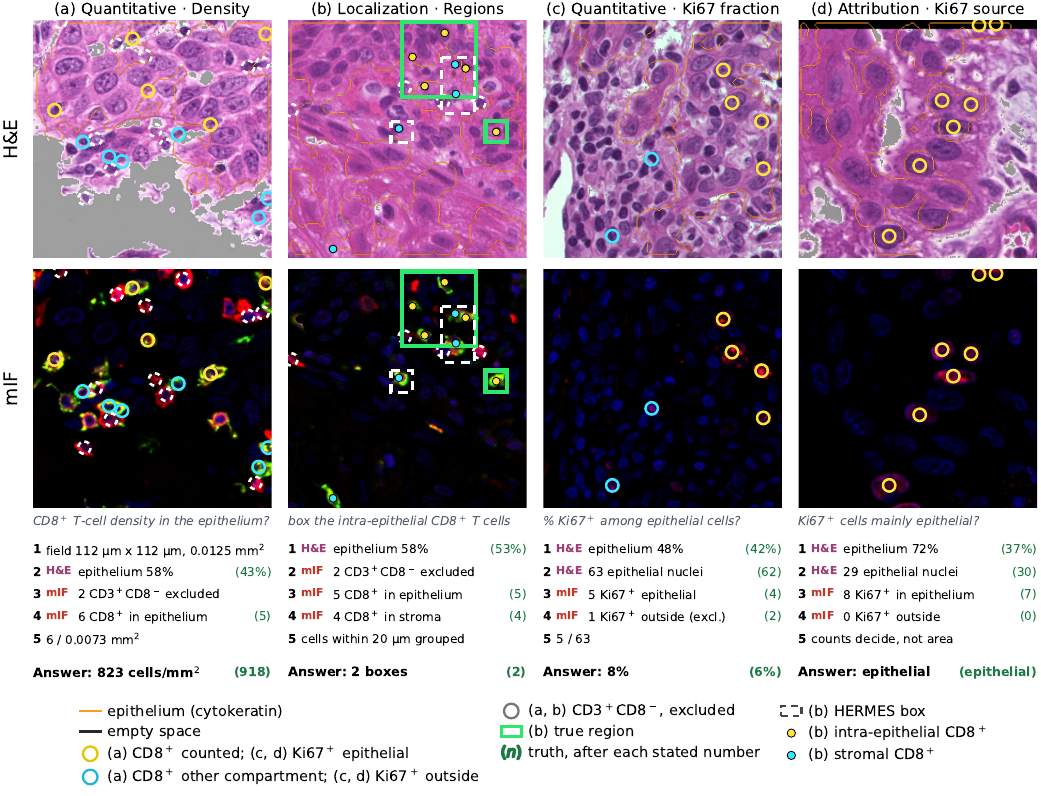}
  \caption{Haist cohort (head and neck cancer, CODEX), example set 3 of 3: one test field per task, laid out as \cref{fig:gt}. \hermes is the model trained on Orion, without retraining. Fields are chosen among correct answers whose stated counts are within one cell of the truth (Ki67 fraction within 3\,pp, density within 15\%), with 30--70\% epithelium and one patient per task and set. Steps without a bracket have no truth in this cohort.}
  \label{fig:haist_ex3}
\end{figure*}

\Cref{fig:schurch_ex1,fig:schurch_ex2,fig:haist_ex1,fig:haist_ex2,fig:haist_ex3} show \hermes's reasoning on test fields of the two external cohorts, in the format of \cref{fig:gt}.

\section{Reward-based results}
\label{sec:suppl_reward}
\Cref{tab:reward} repeats the main comparison with the graded answer reward used in training instead of accuracy. The ranking is unchanged: \hermes (RL) has the highest reward on every task (tied with SFT on Ki67 source) and the highest average.

\begin{table*}[t]
\centering\small
\setlength{\tabcolsep}{4pt}
\providecommand{\hd}[1]{\begin{tabular}[b]{@{}c@{}}#1\end{tabular}}
\caption{Main results on Orion scored with the answer reward used in training ($r_\text{ans}$, \cref{sec:grpo}; graded in $[0, 1]$, $\times 100$) instead of accuracy; same 1,000 questions as \cref{tab:main}, with the answer read from the last \texttt{<answer>} tag as for the accuracy. Average: mean of the four tasks.}
\label{tab:reward}
\begin{tabular}{@{}l ccccc@{}}
\toprule
Model & Density & Regions & \hd{Ki67\\source} & \hd{Ki67\\fraction} & Average \\
\midrule
\rowcolor{gray!12}\multicolumn{6}{c}{\emph{Proprietary MLLMs}} \\
GPT-5.6 Sol & 42.3 & 44.1 & 73.2 & 72.0 & 57.9 \\
Gemini 3.7 Flash & 44.4 & 44.7 & 85.6 & 68.0 & 60.7 \\
Claude Opus 5 & 32.5 & 24.3 & 66.0 & 46.7 & 42.4 \\
\rowcolor{gray!12}\multicolumn{6}{c}{\emph{General MLLMs}} \\
Gemma-4-31B & 30.0 & 43.1 & 80.4 & 61.1 & 53.6 \\
Qwen3.8-27B & 33.2 & 41.4 & 71.2 & 61.3 & 51.8 \\
Ministral-3-14B & 25.4 & 44.4 & 59.6 & 28.2 & 39.4 \\
\rowcolor{gray!12}\multicolumn{6}{c}{\emph{Medical MLLMs}} \\
MedGemma-27B & 11.1 & 6.3 & 50.4 & 18.5 & 21.6 \\
HuatuoGPT-3-27B & 32.5 & 43.4 & 67.6 & 58.1 & 50.4 \\
\rowcolor{gray!12}\multicolumn{6}{c}{\emph{Pathology MLLMs}} \\
Quilt-LLaVA-7B & 3.4 & 0.8 & 2.8 & 2.5 & 2.4 \\
PathGen-LLaVA-13B & 0.8 & 0.6 & 0.0 & 0.0 & 0.4 \\
Patho-R1-7B & 15.8 & 4.8 & 27.6 & 19.1 & 16.8 \\
\midrule
Qwen3.5-4B (base) & 14.0 & 26.2 & 60.4 & 30.7 & 32.8 \\
\hermes (SFT) & 44.2 & 59.4 & \textbf{94.4} & 71.9 & 67.5 \\
\hermes (RL) & \textbf{56.9} & \textbf{66.5} & \textbf{94.4} & \textbf{76.6} & \textbf{73.6} \\
\bottomrule
\end{tabular}
\end{table*}

\section{GRPO without the process reward}
\label{sec:suppl_ablation_process}
To isolate the process reward, we repeat the GRPO stage without it ($r = r_\text{fmt} + r_\text{ans}$, \cref{sec:grpo}), starting from the same SFT model with the same training questions, number of steps, batch size and learning rate.
\Cref{tab:ablation_process} compares it with the SFT model and with \hermes (RL).

Without the process reward, GRPO raises the average accuracy over SFT only from 78.8\% to 80.0\%, whereas with it the average rises to 84.5\%; the gap is largest on Density (66.4 vs.\ 76.0) and Ki67 fraction (81.6 vs.\ 86.4).
During training, the answer reward stays flat without the process reward (0.71 over the first 40 steps, 0.70 over the last 40), while with it the answer reward rises from 0.72 to 0.74.

The reasoning changes more than the answers.
Without the process reward, the share of correct intermediate quantities falls below that of the SFT model (37.2\% vs.\ 58.4\%), whereas with it the share rises to 66.2\%.
The model also leaves the guideline steps.
On the two Ki67 tasks, 96--99\% of its outputs follow a new sequence of steps that counts nuclei on the Hoechst channel of the mIF image rather than the epithelial nuclei on H\&E, and the step that should name the image it reads names the marker instead; only 1.2\% of its epithelial nucleus counts on Ki67 source are correct, against 30\% for \hermes (RL).
It also stops stating some quantities: on Regions it reports a nucleus count in place of the epithelial share, and on Ki67 fraction it rarely states the Ki67$^+$ nuclei outside the epithelium, so only 83.7\% of the needed quantities are stated.
Rewarding only the final answer thus leaves the reasoning free to drift from the guidelines, while the process reward keeps every step on the guideline and checkable against the truth.

\begin{table*}[t]
\centering\small
\setlength{\tabcolsep}{3.5pt}
\providecommand{\hd}[1]{\begin{tabular}[b]{@{}c@{}}#1\end{tabular}}
\caption{GRPO with (\hermes (RL)) and without the process reward $r_\text{proc}$, from the same SFT model and with the same data, steps and hyperparameters (Orion, same 1,000 questions as \cref{tab:main}). Answer accuracy as in \cref{tab:main}. Intermediate quantities: share of the four quantities a task needs that are correct (unstated counts as wrong), extracted by the same LLM judge for all three rows as for the baselines in \cref{fig:iq}; Stated: share of the needed quantities that are stated.}
\label{tab:ablation_process}
\begin{tabular}{@{}l ccccc ccccc c@{}}
\toprule
 & \multicolumn{5}{c}{Answer accuracy} & \multicolumn{6}{c}{Intermediate quantities} \\
\cmidrule(lr){2-6}\cmidrule(lr){7-12}
Model & Density & Regions & \hd{Ki67\\source} & \hd{Ki67\\fraction} & Avg. & Density & Regions & \hd{Ki67\\source} & \hd{Ki67\\fraction} & Correct & Stated \\
\midrule
\hermes (SFT) & 65.2 & 74.5 & \textbf{94.4} & 81.2 & 78.8 & 66.1 & 62.0 & 57.1 & 48.5 & 58.4 & \textbf{100.0} \\
GRPO without $r_\text{proc}$ & 66.4 & 79.4 & 92.4 & 81.6 & 80.0 & 59.2 & 42.6 & 24.1 & 22.8 & 37.2 & 83.7 \\
\hermes (RL) & \textbf{76.0} & \textbf{81.3} & \textbf{94.4} & \textbf{86.4} & \textbf{84.5} & \textbf{75.7} & \textbf{67.8} & \textbf{64.2} & \textbf{57.0} & \textbf{66.2} & \textbf{100.0} \\
\bottomrule
\end{tabular}
\end{table*}

The same holds on the external cohorts without retraining (\cref{tab:ablation_ext}): GRPO without the process reward raises the average accuracy over SFT by 2.4 points on Sch\"urch and 1.1 points on Haist, while \hermes (RL) raises it by 5.5 and 3.0 points.

\begin{table}[t]
\centering\small
\caption{Average answer accuracy (\%) with and without the process reward, on Orion and, without retraining, on the external cohorts (questions as in \cref{tab:main,tab:schurch}).}
\label{tab:ablation_ext}
\begin{tabular}{@{}l ccc@{}}
\toprule
Model & Orion & Sch\"urch & Haist \\
\midrule
\hermes (SFT) & 78.8 & 70.9 & 73.7 \\
GRPO without $r_\text{proc}$ & 80.0 & 73.3 & 74.8 \\
\hermes (RL) & \textbf{84.5} & \textbf{76.4} & \textbf{76.7} \\
\bottomrule
\end{tabular}
\end{table}

\section{Task-specific foundation-model baseline}
\label{sec:suppl_fmbase}
To compare with a model trained specifically for the four tasks on the same inputs, we encode the two images that \hermes receives for each question with frozen pathology foundation models: Prov-GigaPath~\cite{xu2024gigapath} for the H\&E image and KRONOS2~\cite{shaban2025foundation} for the mIF image, split back into the markers it shows (Hoechst with CD3e and CD8a, or Hoechst with Ki67).
On these features we train one MLP head per task on the \hermes training questions (density as a regression of the log density, Ki67 fraction as a regression of the percentage, Ki67 source and infiltrate present vs.\ none as binary classification), select the epoch on the Orion validation set and average three seeds.
The heads output a readout only, without boxes or intermediate quantities.

On Orion, the baseline reaches an average accuracy of 80.4\%, above every MLLM baseline (Gemini 3.7 Flash: 74.7\%) and below \hermes (84.5\%; \cref{tab:fmbase}); it is the most accurate model on Ki67 source (98.0 vs.\ 94.4), while \hermes is ahead on Density (76.0 vs.\ 67.1) and Regions (81.3 vs.\ 72.1).
Applied without retraining to the external cohorts, its average falls to 59.6\% on Sch\"urch and 43.5\% on Haist, below Gemini 3.7 Flash, whereas \hermes keeps 76.4\% and 76.7\%.

\begin{table*}[t]
\centering\small
\setlength{\tabcolsep}{3.5pt}
\providecommand{\hd}[1]{\begin{tabular}[b]{@{}c@{}}#1\end{tabular}}
\caption{Task-specific foundation-model baseline with the same inputs: frozen Prov-GigaPath (H\&E) and KRONOS2 (mIF, the markers \hermes sees) on the two images \hermes receives, with one MLP head per task trained on the \hermes training questions (mean of three seeds). The Orion-trained heads are applied to the external cohorts without retraining, as \hermes is. Answer accuracy and question subsets as in \cref{tab:main,tab:schurch}; Haist: average over the four tasks.}
\label{tab:fmbase}
\begin{tabular}{@{}l ccccc ccccc c@{}}
\toprule
 & \multicolumn{5}{c}{Orion} & \multicolumn{5}{c}{Sch\"urch} & Haist \\
\cmidrule(lr){2-6}\cmidrule(lr){7-11}\cmidrule(lr){12-12}
Model & Density & Regions & \hd{Ki67\\source} & \hd{Ki67\\fraction} & Avg. & Density & Regions & \hd{Ki67\\source} & \hd{Ki67\\fraction} & Avg. & Avg. \\
\midrule
\hermes (RL) & \textbf{76.0} & \textbf{81.3} & 94.4 & \textbf{86.4} & \textbf{84.5} & \textbf{69.6} & \textbf{67.7} & \textbf{92.4} & \textbf{76.0} & \textbf{76.4} & \textbf{76.7} \\
GigaPath + KRONOS2, task heads & 67.1 & 72.1 & \textbf{98.0} & 84.3 & 80.4 & 66.5 & 50.5 & 69.6 & 51.6 & 59.6 & 43.5 \\
Gemini 3.7 Flash & 68.0 & 69.9 & 85.6 & 75.2 & 74.7 & 66.4 & 58.1 & 82.8 & 62.0 & 67.3 & 72.5 \\
\bottomrule
\end{tabular}
\end{table*}

\section{Patient-level analysis}
\label{sec:suppl_patient}
\begin{figure*}[t]
  \centering
  \includegraphics[width=\linewidth]{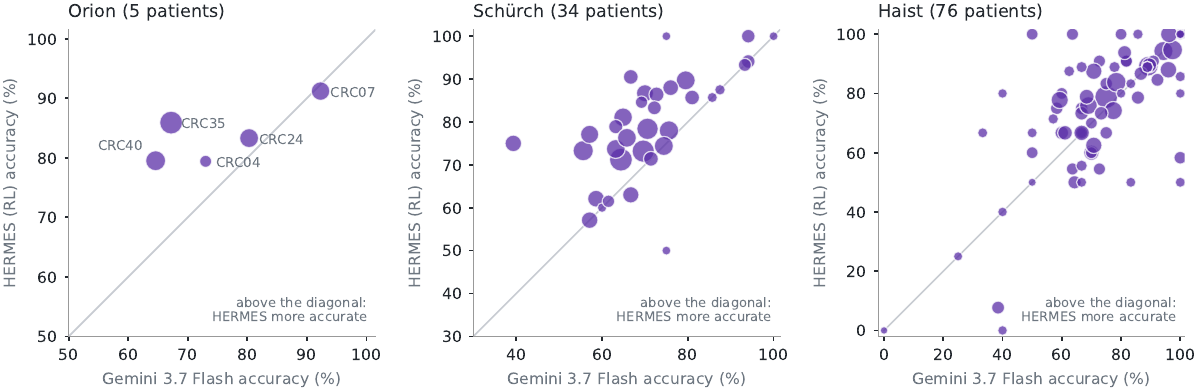}
  \caption{Accuracy per patient, \hermes (RL) against Gemini 3.7 Flash, on all of a patient's questions (Regions scored as the present / none call); marker area: number of questions. Points above the diagonal are patients in which \hermes is more accurate.}
  \label{fig:patient}
\end{figure*}

Fields from the same patient are not independent, so we also compare \hermes (RL) with the strongest baseline, Gemini 3.7 Flash, at the patient level.
We resample patients with replacement, each with all of their questions, recompute the average accuracy as in \cref{tab:main,tab:schurch} and repeat this 10,000 times (\cref{tab:patient}).
The difference stays above zero in every resample on Orion (9.9 points) and Sch\"urch (9.1 points), and in 99.2\% of resamples on Haist (4.2 points).
Per patient (\cref{fig:patient}), \hermes is more accurate in four of the five Orion test patients; in the fifth (CRC07) both models exceed 91\% and differ by 1.1 points.
On Sch\"urch (the 874-question subset covers 34 of the 35 patients), \hermes is more accurate in 22 of 34 patients, less accurate in 2 and equally accurate in 10; on Haist, it is more accurate in 30 of 76 patients, less accurate in 21 and equally accurate in 25.

\begin{table*}[t]
\centering\small
\setlength{\tabcolsep}{5pt}
\caption{Patient-level analysis of the average accuracy (\%): bootstrap over patients (each resample draws patients with all of their questions; 10,000 resamples; 95\% percentile intervals), on the questions of \cref{tab:main,tab:schurch} (Haist: average over the four tasks). W / L / T: patients in which \hermes (RL) is more / less / equally accurate than Gemini 3.7 Flash (\cref{fig:patient}).}
\label{tab:patient}
\begin{tabular}{@{}l cc ccc c@{}}
\toprule
Cohort & Patients & Questions & \hermes (RL) & Gemini 3.7 Flash & Difference & W / L / T \\
\midrule
Orion & 5 & 1,000 & 84.5 [81.8, 86.8] & 74.7 [67.9, 83.2] & 9.9 [2.8, 17.3] & 4 / 1 / 0 \\
Sch\"urch & 34 & 874 & 76.4 [72.9, 80.3] & 67.3 [63.5, 71.3] & 9.1 [6.3, 12.0] & 22 / 2 / 10 \\
Haist & 76 & 1,000 & 76.7 [73.1, 80.2] & 72.5 [69.2, 75.9] & 4.2 [0.9, 7.6] & 30 / 21 / 25 \\
\bottomrule
\end{tabular}
\end{table*}

\section{Visual counterfactuals}
\label{sec:suppl_cf}
To test whether the stated counts are read from the mIF image, we edit the image at known cells and compare the reasoning before and after the edit.
For held-out Orion fields (300 Density, 200 Regions questions), the CD3e and CD8a signal of chosen CD8$^+$ T cells is replaced by the local background of the raw mIF crop, the image is rendered with the unchanged pipeline, and the truth is recomputed with the rules that generate the questions; the H\&E image is not changed.
The edits remove one, three or all CD8$^+$ T cells of the asked compartment, or apply the same edit at three tissue points without a T cell nearby (sham).
Gemma-4-31B and Qwen3.8-27B receive the same images and are asked to report the quantities as JSON, since their free-form reasoning states few of them (\cref{tab:iq}).

\Cref{tab:cf} shows that the counts stated by \hermes follow the edits: removing three CD8$^+$ cells lowers its count in 89.7\% of fields, against 69.6\% and 54.4\% for the baselines, and removing all of them lowers it in 98.3\%.
The sham edit leaves its count and answer unchanged in 73.3\% of fields, against 5.1\% and 16.4\%, so the baselines' counts change even when the cells do not.
On Regions, removing every intra-epithelial CD8$^+$ cell turns the answer of \hermes into ``none'' in 61.0\% of fields (20.9\% and 40.4\%), and after a sham edit it still reports the infiltrate in every field.

\begin{table}[t]
\centering\footnotesize
\setlength{\tabcolsep}{3pt}
\providecommand{\hd}[1]{\begin{tabular}[b]{@{}c@{}}#1\end{tabular}}
\caption{Visual counterfactuals on Orion test fields (300 Density, 200 Regions; \%). The mIF image is edited at known cells and the stated count is compared before and after the edit. Delete 1 / 3 / all: CD3e and CD8a removed from one, three or all CD8$^+$ T cells of the asked compartment (Regions: of the epithelium). Sham: the same edit at three tissue points without a T cell (unchanged: count and answer). Correct: the stated change is within $\max(1, 0.25\,|\Delta|)$ of the true change $\Delta$. Shares over the pairs whose outputs before and after the edit can be read (Density: the count is stated; last row).}
\label{tab:cf}
\begin{tabular}{@{}>{\raggedright\arraybackslash}p{4.05cm}ccc@{}}
\toprule
Edit and outcome & \hermes & \hd{Gemma-\\4-31B} & \hd{Qwen3.8-\\27B} \\
\midrule
\multicolumn{4}{@{}l}{\emph{Density}} \\
Delete 1: change correct & \textbf{93.0} & 50.2 & 50.0 \\
Delete 3: count drops & \textbf{89.7} & 69.6 & 54.4 \\
Delete 3: change correct & \textbf{65.3} & 36.1 & 27.2 \\
Delete all: count drops & \textbf{98.3} & 90.5 & 74.8 \\
Sham: unchanged & \textbf{73.3} & 5.1 & 16.4 \\
Sham: count drops & \textbf{13.0} & 39.5 & 33.1 \\
\multicolumn{4}{@{}l}{\emph{Regions}} \\
Delete all: answer ``none'' & \textbf{61.0} & 20.9 & 40.4 \\
Sham: infiltrate still reported & \textbf{100.0} & 94.4 & 82.6 \\
\midrule
Readable output pairs & \textbf{100.0} & 73.9 & 94.4 \\
\bottomrule
\end{tabular}
\end{table}

\section{Box-level results on Regions}
\label{sec:suppl_regionbox}
Because the regions group intra-epithelial CD8$^+$ cells by a distance threshold, a field can be boxed correctly with fewer or more boxes than the truth; we therefore count a match at a low IoU ($\geq 0.1$) and also report a stricter threshold ($\geq 0.3$).
\Cref{tab:regionbox} shows that \hermes (RL) reaches a box F1 of 60.3\% and 46.0\%, against 48.9\% (GPT-5.6 Sol) and 38.3\% (Gemini 3.7 Flash) for the strongest baselines, and 46.3\% and 25.1\% after SFT.

\begin{table}[t]
\centering\small
\caption{Box-level results on Regions (Orion, same questions as \cref{tab:main}; \%). True boxes are the regions of \cref{sec:tasks}. Box F1: predicted and true boxes matched one-to-one by IoU, a match counting at IoU $\geq 0.1$ or $\geq 0.3$, pooled over all fields (boxes drawn on fields without infiltrate are false positives).}
\label{tab:regionbox}
\begin{tabular}{@{}l cc@{}}
\toprule
 & \multicolumn{2}{c}{Box F1} \\
\cmidrule(lr){2-3}
Model & IoU $\geq 0.1$ & IoU $\geq 0.3$ \\
\midrule
\multicolumn{3}{@{}l}{\emph{Proprietary MLLMs}} \\
GPT-5.6 Sol & 48.9 & 35.4 \\
Gemini 3.7 Flash & 46.9 & 38.3 \\
Claude Opus 5 & 30.2 & 15.1 \\
\multicolumn{3}{@{}l}{\emph{General MLLMs}} \\
Gemma-4-31B & 48.7 & 32.0 \\
Qwen3.8-27B & 35.1 & 17.7 \\
Ministral-3-14B & 3.5 & 0.0 \\
\multicolumn{3}{@{}l}{\emph{Medical MLLMs}} \\
MedGemma-27B & 8.9 & 3.8 \\
HuatuoGPT-3-27B & 38.9 & 19.6 \\
\multicolumn{3}{@{}l}{\emph{Pathology MLLMs}} \\
Quilt-LLaVA-7B & 0.7 & 0.0 \\
PathGen-LLaVA-13B & 0.7 & 0.0 \\
Patho-R1-7B & 0.0 & 0.0 \\
\midrule
Qwen3.5-4B (base) & 12.0 & 4.3 \\
\hermes (SFT) & 46.3 & 25.1 \\
\hermes (RL) & \textbf{60.3} & \textbf{46.0} \\
\bottomrule
\end{tabular}
\end{table}

\section{GRPO seeds}
\label{sec:suppl_seeds}
We repeat the GRPO stage with two further seeds, starting from the same SFT model with the same training questions, rewards and hyperparameters (\cref{tab:seeds}).
Every seed has a higher average accuracy than the strongest baseline on all three cohorts, and the share of correct intermediate quantities varies by less than 1.5 points across seeds.

\begin{table}[t]
\centering\footnotesize
\setlength{\tabcolsep}{2.5pt}
\caption{\hermes (RL) trained with three GRPO seeds from the same SFT model, with the same data and hyperparameters. Average answer accuracy (\%) on the questions of \cref{tab:main,tab:schurch} and share of correct intermediate quantities on Orion (IQ, \%; \cref{tab:iq}). Seed 0 is the model reported in the paper. Strongest baseline: Gemini 3.7 Flash for accuracy, Gemma-4-31B for IQ (proprietary models hide their reasoning).}
\label{tab:seeds}
\begin{tabular}{@{}l cccc@{}}
\toprule
GRPO seed & Orion & Sch\"urch & Haist & IQ \\
\midrule
Seed 0 (reported) & 84.5 & 76.4 & 76.7 & 66.2 \\
Seed 1 & 83.4 & 77.2 & 77.6 & 66.8 \\
Seed 2 & 82.7 & 74.1 & 75.4 & 65.4 \\
\midrule
Mean $\pm$ SD & 83.5 $\pm$ 0.9 & 75.9 $\pm$ 1.6 & 76.6 $\pm$ 1.1 & 66.1 $\pm$ 0.7 \\
Strongest baseline & 74.7 & 67.3 & 72.5 & 16.4 \\
\bottomrule
\end{tabular}
\end{table}

\end{document}